%% file: ms.tex
\documentclass[review]{elsarticle}
\usepackage{dcolumn}
\usepackage{booktabs}
\usepackage{pbox}
\usepackage{multirow}
\usepackage{subfig}
\usepackage{graphicx}
\usepackage{amssymb,calc,url,enumerate}
\usepackage{amsthm}
\usepackage{dsfont}
\usepackage{bm}
\usepackage{booktabs}
\usepackage[centertags]{amsmath}
\usepackage[ruled, linesnumbered]{algorithm2e}
\usepackage{lineno}
\usepackage[hidelinks]{hyperref}
\usepackage{wrapfig}
\usepackage{xcolor}
\usepackage{cancel}
\usepackage{import}
\usepackage[utf8x]{inputenc}

\newcommand{\R}{\mathds{R}}
\newcommand{\I}{\bm{I}}

\newcommand{\So}{\mathcal{S}}
\newcommand{\boldZero}{{\bm{0}}}
\newcommand{\transpose}{^\mathsf{T}}

\newcommand{\inv}[1]{#1^{-1}}
\newcommand{\norm}[1] {\parallel #1 \parallel}
\newcommand{\innerprod}[1] {\langle #1 \rangle}

\newcommand{\bbeta}{{\bm{\beta}}}

\newcommand{\nnu}{{\bm{\nu}}}

\newcommand{\ab}{{\bm{a}}}

\newcommand{\oo}{{\bm{o}}}
\newcommand{\tb}{{\bm{t}}}

\newcommand{\vv}{{\bm{v}}}
\newcommand{\xx}{{\bm{x}}}

\newcommand{\Ab}{{\bm{A}}}
\newcommand{\BB}{{\bm{B}}}
\newcommand{\CC}{{\bm{C}}}
\newcommand{\DD}{{\bm{D}}}
\newcommand{\EE}{{\bm{E}}}

\newcommand{\HH}{{\bm{H}}}
\newcommand{\JJ}{{\bm{J}}}
\newcommand{\KK}{{\bm{K}}}
\newcommand{\LL}{{\bm{L}}}

\newcommand{\TT}{{\bm{T}}}
\newcommand{\UU}{{\bm{U}}}

\newcommand{\XX}{{\bm{X}}}
\newcommand{\YY}{{\bm{Y}}}
\newcommand{\ZZ}{{\bm{Z}}}
\newcommand{\Ntil}{{\widetilde{N}}}

\DeclareMathOperator*{\argmin}{arg\,min}

\newcommand{\equref}[1]{Eq.~\ref{#1}}
\newcommand{\figref}[1]{Fig.~\ref{#1}}
\newcommand{\secref}[1]{Section~\ref{#1}}
\newcommand{\tabref}[1]{Table~\ref{#1}}
\newcommand{\algref}[1]{Algorithm~\ref{#1}}

\modulolinenumbers[5]
\journal{Expert Systems With Applications}

\biboptions{authoryear}

\begin{document}

\begin{frontmatter}

\title{Outlier Robust Extreme Learning Machine for Multi-Target Regression}


\author[ufes]{Bruno L\'{e}gora Souza da Silva\corref{correspauthor}}
\cortext[correspauthor]{Corresponding author}
\ead{bruno.l.silva@aluno.ufes.br}

\author[ufes]{Fernando Kentaro Inaba}
\ead{fkinaba@gmail.com}

\author[ufes]{Evandro Ottoni Teatini Salles}
\ead{evandro@ele.ufes.br}

\author[ufes]{Patrick Marques Ciarelli}
\ead{patrick.ciarelli@ufes.br}

\address[ufes]{Federal University of Esp\'{i}rito Santo, Vit\'{o}ria, Brazil}





\begin{abstract}

The popularity of algorithms based on Extreme Learning Machine (ELM), which can be used to train Single Layer Feedforward Neural Networks (SLFN), has increased in the past years. They have been successfully applied to a wide range of classification and regression tasks. The most commonly used methods are the ones based on minimizing the $\ell_2$ norm of the error, which is not suitable to deal with outliers, essentially in regression tasks. The use of $\ell_1$ norm was proposed in Outlier Robust ELM (OR-ELM), which is defined to one-dimensional outputs. In this paper, we generalize OR-ELM to deal with multi-target regression problems, using the error $\ell_{2,1}$ norm and the Elastic Net theory, which can result in a more sparse network, resulting in our method, Generalized Outlier Robust ELM (GOR-ELM). We use Alternating Direction Method of Multipliers (ADMM) to solve the resulting optimization problem. An incremental version of GOR-ELM is also proposed. We chose 15 public real-world multi-target regression datasets to test our methods. Our conducted experiments show that they are statistically better than other ELM-based techniques, when considering data contaminated with outliers, and equivalent to them, otherwise.

\end{abstract}


\end{frontmatter}

\input{sec/intro.tex}

\input{sec/relatedwork.tex}

\input{sec/method.tex}

\input{sec/igor.tex}

\input{sec/experiments.tex}

\input{sec/conclusions.tex}

\section*{Acknowledgments}
The authors wish to acknowledge the support of the Brazilian Federal Agency for Support and Evaluation of Graduate Education (CAPES) through Grants 88881.062156/2014-01 and 1753509. F. Inaba was supported by a scholarship from the Brazilian Council for Development of Science and Technology (CNPq). P. Ciarelli thanks the partial funding of his research work provided by CNPq through grant 312032/2015-3. 
The work of E. O. T. Salles was support by Fundação de Amparo \`{a} Pesquisa e Inova\c{c}\~{a}o do Esp\'{i}rito Santo (FAPES) under Grant 244/2016.


\section*{References}

\bibliography{mybibfile}

\end{document}

%% file: sec/intro.tex
\section{Introduction}\label{sec:intro}

\begin{figure}[htb]
	\begin{center}
		\resizebox{0.7\linewidth}{!}{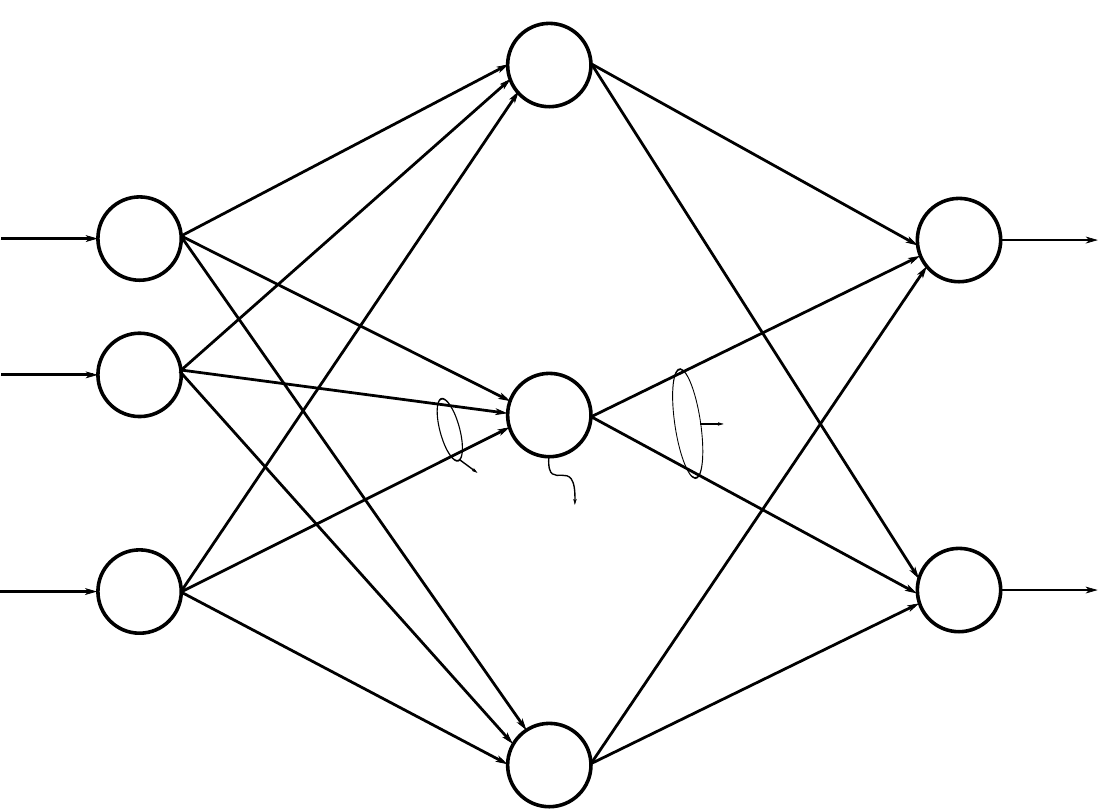}
	\end{center}
	\caption{Illustration of an SLFN architecture with $n$ inputs, $\Ntil$ hidden nodes and $m$ outputs \citep{Inaba2018}.}
	\label{fig:ELM}
\end{figure}

In the past years, a fast algorithm called Extreme Learning Machine (ELM) was proposed by \citet{Huang2004,Huang2006}. It is used to train Single Layer Feedforward Networks (SLFN), as shown in \figref{fig:ELM}, where part of the network parameters ($\ab_i$ and $\nu_i$) are randomly generated, and the remaining ($\bbeta_i$) are found using labeled data and a closed-form solution. 

Due to its simplicity and speed, ELM gained popularity and has been applied in a wide range of applications such as computer vision and time series analysis \citep{Huang2015}, achieving good performances, generally better than classifiers such as Support Vector Machines (SVMs) \citep{Huang2015}.

Variants of ELM were also proposed, turning possible to deal with sequential data (where new samples arrive over time) using Online Sequential ELM (OS-ELM) \citep{Liang2006}, or increasing the SLFN hidden node number, using Incremental ELM (I-ELM) \citep{Huang2006a}. 

In ELM and OS-ELM, the number of nodes in the hidden layer needs to be well-chosen to avoid overfitting and underfitting \citep{Deng2009}. To overcome this problem, an ELM-based algorithm using ridge regression was proposed by \citet{Deng2009}. Although it achieves good results, the resulting network is dense and might suffer from memory and computing limitations.

\citet{Martinez-Martinez2011} extended \citet{Deng2009} work, proposing an algorithm named Regularized ELM (R-ELM), which can select an architecture based on the problem. By using the Elastic Net theory, R-ELM can prune some hidden nodes when dealing with a one-dimensional output.

The aforementioned methods were defined considering only one output node, with optimization problems minimizing vector norms, but they can be adapted to multi-dimensional tasks by considering each output separately. To deal with these tasks, Generalized Regularized ELM (GR-ELM) was proposed by \citet{Inaba2018}, which generalizes R-ELM, using matrix norms in its objective function, replacing $\ell_2$ and $\ell_1$ vector norms by Frobenius and $\ell_{2,1}$ matrix norms, respectively. 

One common characteristic of these methods is the use of $\ell_2$ or Frobenius norm of the prediction error in each objective function. As pointed by \citet{Zhang2015}, these approaches have some drawbacks when the application suffers from outliers, which is common in real-world tasks, since the model can be biased towards them.

To deal with outliers, Outlier Robust ELM (OR-ELM) was proposed by \citet{Zhang2015}, which considers the $\ell_1$ norm of the prediction error in its objective function, achieving better results in the presence of outliers in regression tasks, when compared with other algorithms. However, OR-ELM was also defined only to one-dimensional outputs. 

In this paper, we generalize the OR-ELM to multi-target regression problems. We use the same model considered in GR-ELM, replacing Frobenius norm by $\ell_{2,1}$ norm  of the prediction error, which can be interpreted as an extension of the $\ell_1$ vector norm. When considering outputs with only one dimension and using only the ridge penalty, our method is the same as OR-ELM. We use a three-block Alternating Direction Method of Multipliers (ADMM) \citep{chen2016direct} to solve our optimization problem, which is a simple but powerful algorithm. We also propose an incremental version of this method, which can increase the number of nodes efficiently if the desired error was not obtained.

Our methods were tested in 15 public real-world datasets, which were contaminated with outliers to verify the robustness of them, using the average relative root mean square error (aRRMSE) as metric. We compared our methods with similar algorithms based on ELM.

The paper is organized as follows. In \secref{sec:RelWork}, we review some ELM-based algorithms. We describe the proposed methods in \secref{sec:PropMeth}. Experiments and results are presented in \secref{sec:Exper}, and \secref{sec:Concl} concludes the  paper.

%% file: figs/ELM.pdf_tex
\begingroup%
  \makeatletter%
  \providecommand\color[2][]{%
    \errmessage{(Inkscape) Color is used for the text in Inkscape, but the package 'color.sty' is not loaded}%
    \renewcommand\color[2][]{}%
  }%
  \providecommand\transparent[1]{%
    \errmessage{(Inkscape) Transparency is used (non-zero) for the text in Inkscape, but the package 'transparent.sty' is not loaded}%
    \renewcommand\transparent[1]{}%
  }%
  \providecommand\rotatebox[2]{#2}%
  \ifx\svgwidth\undefined%
    \setlength{\unitlength}{316.19602051bp}%
    \ifx\svgscale\undefined%
      \relax%
    \else%
      \setlength{\unitlength}{\unitlength * \real{\svgscale}}%
    \fi%
  \else%
    \setlength{\unitlength}{\svgwidth}%
  \fi%
  \global\let\svgwidth\undefined%
  \global\let\svgscale\undefined%
  \makeatother%
  \begin{picture}(1,0.73644279)%
    \put(0,0){\includegraphics[width=\unitlength]{figs/ELM.pdf}}%
    \put(-0.00201286,0.52303684){\color[rgb]{0,0,0}\makebox(0,0)[lb]{\smash{$1$}}}%
    \put(-0.00201286,0.4098159){\color[rgb]{0,0,0}\makebox(0,0)[lb]{\smash{$x_{k_1}$}}}%
    \put(-0.00201286,0.21183748){\color[rgb]{0,0,0}\makebox(0,0)[lb]{\smash{$x_{k_n}$}}}%
    \put(0.13034951,0.26696172){\color[rgb]{0,0,0}\rotatebox{90}{\makebox(0,0)[lb]{\smash{.}}}}%
    \put(0.13034951,0.28985385){\color[rgb]{0,0,0}\rotatebox{90}{\makebox(0,0)[lb]{\smash{.}}}}%
    \put(0.13034951,0.31274598){\color[rgb]{0,0,0}\rotatebox{90}{\makebox(0,0)[lb]{\smash{.}}}}%
    \put(0.50353578,0.49116204){\color[rgb]{0,0,0}\rotatebox{90}{\makebox(0,0)[lb]{\smash{.}}}}%
    \put(0.50353578,0.51405416){\color[rgb]{0,0,0}\rotatebox{90}{\makebox(0,0)[lb]{\smash{.}}}}%
    \put(0.50353578,0.53694629){\color[rgb]{0,0,0}\rotatebox{90}{\makebox(0,0)[lb]{\smash{.}}}}%
    \put(0.50353578,0.16997778){\color[rgb]{0,0,0}\rotatebox{90}{\makebox(0,0)[lb]{\smash{.}}}}%
    \put(0.50353578,0.19286991){\color[rgb]{0,0,0}\rotatebox{90}{\makebox(0,0)[lb]{\smash{.}}}}%
    \put(0.50353578,0.21576204){\color[rgb]{0,0,0}\rotatebox{90}{\makebox(0,0)[lb]{\smash{.}}}}%
    \put(0.87672198,0.33227128){\color[rgb]{0,0,0}\rotatebox{90}{\makebox(0,0)[lb]{\smash{.}}}}%
    \put(0.87672198,0.35516341){\color[rgb]{0,0,0}\rotatebox{90}{\makebox(0,0)[lb]{\smash{.}}}}%
    \put(0.87672198,0.37805554){\color[rgb]{0,0,0}\rotatebox{90}{\makebox(0,0)[lb]{\smash{.}}}}%
    \put(0.43726461,0.4007988){\color[rgb]{0,0,0}\makebox(0,0)[lb]{\smash{$\nu_i$}}}%
    \put(0.43286168,0.29028339){\color[rgb]{0,0,0}\makebox(0,0)[lb]{\smash{$\ab_i$}}}%
    \put(0.51591528,0.2518303){\color[rgb]{0,0,0}\makebox(0,0)[lb]{\smash{$h(\xx_k;\ab_i,\nu_i)$}}}%
    \put(0.06767954,0.1367326){\color[rgb]{0,0,0}\makebox(0,0)[lb]{\smash{$\xx_k$}}}%
    \put(0.90166205,0.1367326){\color[rgb]{0,0,0}\makebox(0,0)[lb]{\smash{$\oo_k$}}}%
    \put(0.9563575,0.53358604){\color[rgb]{0,0,0}\makebox(0,0)[lb]{\smash{$o_{k_1}$}}}%
    \put(0.95208305,0.21424348){\color[rgb]{0,0,0}\makebox(0,0)[lb]{\smash{$o_{k_m}$}}}%
    \put(0.49198433,0.66672211){\color[rgb]{0,0,0}\makebox(0,0)[lb]{\smash{$1$}}}%
    \put(0.49376731,0.35094928){\color[rgb]{0,0,0}\makebox(0,0)[lb]{\smash{$i$}}}%
    \put(0.48572273,0.02683559){\color[rgb]{0,0,0}\makebox(0,0)[lb]{\smash{$\Ntil$}}}%
    \put(0.66072542,0.3428239){\color[rgb]{0,0,0}\makebox(0,0)[lb]{\smash{$\bbeta_i$}}}%
  \end{picture}%
\endgroup%

%% file: sec/relatedwork.tex
\section{Related Work}\label{sec:RelWork}

In this section, we review Extreme Learning Machine and some of its variants, in specific regularized and incremental ones.

\subsection{ELM}

An extremely fast algorithm to train an SLFN with $\Ntil$ hidden nodes was proposed in \citet{Huang2004}. This algorithm was called ELM and considers a dataset with $N$ distinct and labeled samples $(\xx_j,t_j)$, where $\xx_j = [x_{j1},x_{j2},\ldots,x_{jn}] \in \R^n$ and $t_j \in \R$. The SLFN estimation of $t_j$ is modeled as
\begin{equation}
\label{eq:slfnoutput}
\centering
\hat{t}_j = \sum_{i=1}^{\Ntil} \beta_ih(\ab_i\cdot\xx_j + \nu_i)
\end{equation}

\noindent where $\ab_i = [a_{i1},a_{i2},\ldots,a_{in}]\transpose$ is the weight vector connecting the $i$-th hidden node and the input nodes. $\bbeta = [\beta_{1},\beta_{2},\ldots,\beta_{\Ntil}]\transpose$ is the weight vector connecting the hidden nodes and the output node, $\nu_i$ is the bias of the $i$-th hidden node, and $h(\cdot)$ is the activation function of the SLFN.

Assuming that an SLFN with $\Ntil$ nodes can approximate all $N$ samples with zero error, i.e, $\hat{t}_j = t_j, j=1,\ldots,N$, we can rewrite \equref{eq:slfnoutput} as:
\begin{equation}
\label{eq:elm_hbt}
\HH\bbeta = \tb
\end{equation}
where $\tb = [t_1,\ldots,t_N]\transpose$ and
\begin{equation}
\centering
\HH(\Ab,\nnu,\XX) = \begin{bmatrix}
h(\ab_1\cdot\xx_1 + \nu_1) & \cdots & h(\ab_\Ntil\cdot\xx_1 + \nu_\Ntil)\\
\vdots & \ddots & \vdots\\
h(\ab_1\cdot\xx_N + \nu_1) & \ldots & h(\ab_\Ntil\cdot\xx_N + \nu_\Ntil)
\end{bmatrix}_{N \times \Ntil},
\label{eq:elm_hmatrix}
\end{equation}
where $\Ab = [\ab_1,\ldots,\ab_\Ntil]$, $\nnu = [\nu_1,\ldots,\nu_\Ntil]\transpose$ and $\XX = [\xx_1\transpose,\ldots,\xx_N\transpose]\transpose$.

The ELM solution, $\hat{\bbeta}$, is the smallest norm least-square solution of the linear system given by \equref{eq:elm_hbt}, i.e.,
\begin{equation}
\centering
\hat{\bbeta} = \argmin_{\bbeta} \norm{\HH\bbeta - \tb}_2^2,
\label{eq:elm_opt}
\end{equation}
\noindent which has a closed-form solution:
\begin{equation}
\hat{\bbeta}= \HH^\dagger\tb,
\label{eq:elm_sol}
\end{equation}
\noindent where $\HH^\dagger$ is the Moore-Penrose generalized inverse \citep{rao1971generalized} of $\HH$.

\subsection{Regularized ELM}

Although ELM has shown good results in several applications, the right choice of $\Ntil$ must be made in order to obtain a good performance, avoiding overfitting and underfitting. \citet{bartlett1998sample} showed that models whose parameters have smaller norms are capable of achieving better generalization. On this note, \citet{Deng2009} introduced R-ELM for SLFN with sigmoid additive nodes, where an $\ell_2$ norm of $\bbeta$ is added in the optimization problem of \equref{eq:elm_opt}. \citet{Huang2012} extends this work to different types of hidden nodes and activation functions, as well as kernels.

Other types of regularization in the ELM optimization problem were considered by \citet{Martinez-Martinez2011}. Then, R-ELM can be described in a generalized way by
\begin{equation}
\centering
\underset{(\beta_0,\bbeta) \in \R^{\Ntil+1}}{\text{minimize}} \frac{C}{2}\norm{\HH\bbeta + \beta_0 - \tb}_2^2 + \frac{(1-\alpha)}{2}\norm{\bbeta}_2^2 + \alpha\norm{\bbeta}_1,
\label{eq:relm_opt}
\end{equation}
where $C$ and $\beta_0$ are regularization parameters.

The optimization problem of \equref{eq:relm_opt} uses the Elastic Net penalty, which is a trade-off between the ridge regression ($\alpha = 0$), where only the $\ell_2$ norm is considered, and the lasso penalty ($\alpha = 1$), where only the $\ell_1$ norm is considered. Since it uses the $\ell_1$ norm, the Elastic Net method is also capable of reducing the network size, pruning output weights while maintaining generalization.

When considering only the $\ell_2$ penalty ($\alpha = 0$) with $\beta_0 = 0$, as done by \citet{Deng2009} and \citet{Huang2012}, the solution of R-ELM is given by
\begin{equation}
\centering
\hat{\bbeta} = \inv{\left(\HH\transpose\HH + \frac{\I}{C}\right)}\HH\transpose\tb.
\label{eq:relm_solut}
\end{equation}

\subsection{Outlier Robust ELM}

Recently, the $\ell_1$ norm has been applied to improve the performance of methods in the presence of outliers \citep{Zhang2015}. This can be achieved in ELM by using it in the loss function, i.e.,
\begin{equation}
\centering
\hat{\bbeta}_r = \argmin_\bbeta \norm{\HH\bbeta - \tb}_1,
\end{equation}
instead of using the usual $\ell_2$ norm.

Supported on this observation, \citet{Zhang2015} proposed the OR-ELM algorithm, which finds the solution of the following optimization problem:
\begin{equation}
\centering
\hat{\bbeta} = \argmin_{\bbeta}~\tau\norm{\HH\bbeta - \tb}_1 + \frac{1}{2C}\norm{\bbeta}_2^2.
\end{equation}

This optimization problem can be solved by using the Augmented Lagrange Multiplier (ALM) method, as suggested by \citet{Zhang2015}. 

\subsection{Generalized Regularized ELM}

A limitation of ELM, R-ELM and OR-ELM is that they are defined only for one-dimensional outputs. Although we can consider each output separately in multi-dimensional tasks, their solutions are not capable of capturing relations between the multiple outputs. This means that we can only prune a hidden node if and only if its weights are pruned in all outputs simultaneously, which can be difficult.

To capture those relations, matrix norms can be used in the ELM optimization problem. Considering this, the GR-ELM method was proposed by \citet{Inaba2018}. This method considers a optimization problem similar to \equref{eq:relm_opt}, using matrices and replacing $\ell_2$ and $\ell_1$ norms by Frobenius and $\ell_{2,1}$ norms, respectively.

Considering a dataset with multi-dimensional outputs, with $N$ distinct samples $(\xx_i,\tb_i)$, where $\xx_i = [x_{i1},x_{i2},\ldots,x_{in}] \in \R^n$ and $\tb_i = [t_{i1},t_{i2},\ldots,t_{im}] \in \R^m$, we can extend \equref{eq:elm_hbt}: 
\begin{equation}
\label{eq:elm_hbT}
\HH\BB = \TT,
\end{equation}
\noindent where $\TT = [\tb_1\transpose,\tb_2\transpose,\ldots,\tb_N\transpose]\transpose$ and $\BB$ is the weight matrix connecting the SLFN hidden and output layers, i.e., $\BB = [\bbeta_1,\bbeta_2,\ldots,\bbeta_m] \in \R^{\Ntil\times m}$. Then, the optimization problem of GR-ELM is given by

\begin{equation}
\begin{aligned}
& \underset{\BB,\ZZ}{\text{minimize}}
& & \frac{C}{2}\norm{\HH\BB - \TT}_F^2 + \frac{\lambda(1-\alpha)}{2}\norm{\BB}_F^2 + \lambda\alpha\norm{\ZZ}_{2,1}\\
& \text{subject to}
& & \BB - \ZZ = \boldZero, \\
\end{aligned}
\label{eq:GR_ELM_ADMM}
\end{equation}

\noindent where $C$, $\lambda$ and $\alpha$ are regularization parameters, $\norm{\Ab}_F$ is the Frobenius norm of a $\Ab = [\ab_{1,\cdot}\transpose,\ab_{2,\cdot}\transpose,\ldots,\ab_{n,\cdot}\transpose]\transpose$ matrix with $n$ rows and $m$ columns, where $\ab_{i,\cdot}$ is the $i$-th row of $\Ab$, defined as
\begin{equation}
\norm{\Ab}_F ~:=~ \sqrt{\sum_{i=1}^{n}\sum_{j=1}^{m} |a_{ij}|^2} = \sqrt{tr(\Ab\transpose\Ab)},
\end{equation}
\noindent where $tr(\Ab\transpose\Ab)$ is the trace of $\Ab\transpose\Ab$ and  $\norm{\Ab}_{2,1}$ is the $\ell_{2,1}$ norm of $\Ab$, defined as
\begin{equation}
\norm{\Ab}_{2,1} ~:=~ \sum_{i=1}^{n} \norm{\ab_{i,\cdot}}_2 = \sum_{i=1}^{n} \left(\sum_{j=1}^{m} |a_{ij}|^2\right)^{1/2}.
\end{equation}

This optimization problem can be solved by using the Alternating Direction Method of Multipliers (ADMM) \citep{Boyd2010a} method, as suggested by \citet{Inaba2018}. \algref{alg:grelm} summarizes the GR-ELM method, where $\rho$ is an ADMM parameter.

\begin{algorithm}[H]
	\textbf{Require:} Training samples ($\XX$,$\TT$), where $\XX = [\xx_1,\xx_2,\ldots,\xx_N]\transpose$, activation function $h(\cdot)$,  number of hidden nodes $\tilde{N}$, and regularization parameters $\lambda$, $\alpha$, C and $\rho$.\\
	\textbf{Initialization}: $\BB^0$, $\UU^0$, $\ZZ^0$ and $k=0$;\\
		Generate new random weights $\mathbf{A}$ and biases $\mathbf{\nnu}$.\\
		Calculate the output of hidden nodes $\HH$ using \equref{eq:elm_hbt}.\\
		\Repeat{meet stopping criterion}{
			$\mathbf{B}^{k+1} := \arg\min_{\mathbf{B}}~ \frac{C}{2} \norm{\mathbf{HB} - \mathbf{T}}_F^2 + \frac{\lambda(1-\alpha)}{2}\norm{\mathbf{B}}_F^2 + \frac{\rho}{2} \norm{\mathbf{B} - \mathbf{Z}^k + \mathbf{U}^{k}}_F^2$\\
			$\mathbf{Z}^{k+1} := \arg\min_{\mathbf{Z}}~ \frac{\lambda\alpha}{\rho} \norm{\textbf{Z}}_{2,1} + \frac{1}{2} \norm{\mathbf{B}^{k+1} - \mathbf{Z} + \mathbf{U}^{k}}_F^2$\\
			$\mathbf{U}^{k+1} := \mathbf{U}^{k} + \rho\left(\mathbf{B}^{k+1} - \mathbf{Z}^{k+1}\right)$\\
			k = k+1
		}
	\caption{GR-ELM: Generalized Robust Extreme Learning Machine}
	\label{alg:grelm}
\end{algorithm}

\subsection{Incremental ELM}
\label{sec:IELM}
Due to its simplicity and low computational cost, the ELM algorithm was very popular after it was proposed. However, it has two main challenges \citep{GuoruiFeng2009}: reducing its complexity to deal with large datasets and models with many nodes; and choosing the optimal hidden node number (usually a trial and error method is used).

To deal with those problems, \citet{Huang2006} proposed an algorithm named I-ELM to train an SLFN adding one node at a time, reducing the memory cost (since in each iteration, it only works with one node) of training an SLFN. Considering a dataset with one-dimensional targets, the algorithm starts with $\Ntil = 0$ and considers that the residual error is equal to the targets: $\mathbf{e}^{(0)} = \tb$. 

When the $j$-th node is added, its input weights and bias are randomly generated, and then its output of every training sample is calculated as $\HH_j = \left[h(\ab_j\cdot\xx_1 + b_j),\ldots,h(\ab_j\cdot\xx_N + b_j)\right]\transpose$. Finally, the output weight $\beta_{j}$ of the new node is calculated using
\begin{equation}
\label{eq:ielm_outw}
\beta_{j} = \frac{\HH_{j}\transpose \mathbf{e}^{(j-1)}}{\HH_{j}\transpose \HH_{j}},
\end{equation}
and the residual error is updated using
\begin{equation}
\label{eq:ielm_error}
\mathbf{e}^{(j)} = \mathbf{e}^{(j-1)} - \beta_{j}\HH_{j}.
\end{equation}

Following this algorithm, it is possible to achieve good approximations of functions, with good generalization performance \citep{Huang2006}. Furthermore, by using an expected learning accuracy as a stopping criterion, it is possible to find a good number of hidden nodes without using the trial and error approach. 

One disadvantage of the I-ELM algorithm is that only one hidden node can be added at a time. An incremental method that was capable of adding a group of hidden nodes at the same time in an SLFN was proposed by \citet{GuoruiFeng2009}. This method is called Error Minimization ELM (EM-ELM) and takes advantage of the Schur complement \citep{Petersen2007} to update the generalized inverse of the matrix $\HH_k$, which is updated in each $k$-th iteration.

Assuming we already have an SLFN, with one output and $\Ntil$ hidden nodes, trained using the ELM algorithm, we used a $\HH_k$ matrix to calculate the output weights $\bbeta_k$. When $\delta \Ntil$ new nodes are added to the network, its new hidden output matrix $\HH_{k+1}$ can be written as:
\begin{equation}
\centering
\HH_{k+1} = \left[\HH_k, \delta\HH_k\right].
\label{eq:newHemelm}
\end{equation}
\noindent where 
\begin{equation}
\centering
\delta\HH_k = \begin{bmatrix}
h(\ab_{\Ntil+1}\cdot\xx_1 + b_{\Ntil+1}) & \cdots & h(\ab_{\Ntil+\delta\Ntil}\cdot\xx_1 + b_{\Ntil+\delta\Ntil})\\
\vdots & \ddots & \vdots\\
h(\ab_{\Ntil+1}\cdot\xx_N + b_{\Ntil+1}) & \ldots & h(\ab_{\Ntil+\delta\Ntil}\cdot\xx_N + b_{\Ntil+\delta\Ntil})
\end{bmatrix}_{N \times \delta\Ntil}.
\label{eq:newHemelm2}
\end{equation}

According to \citet{GuoruiFeng2009}, it is possible to update the output weights $\bm{\beta}_{k+1}$:
\begin{equation}
\bm{\beta}_{k+1} = \HH_{k+1}^{\dagger}\tb = \begin{bmatrix}
\mathbf{U}_k\\
\mathbf{D}_k
\end{bmatrix}\tb,
\label{eq:emelm_bkp1}
\end{equation}
where $\mathbf{U}_k$ and $\mathbf{D}_k$ are given by
\begin{equation}
\label{eq:emelm_uk}
\mathbf{U}_k = \HH_{k}^{\dagger}\left(\mathbf{I} - \delta\HH_k\transpose\mathbf{D}_k\right),
\end{equation}
where $\mathbf{I}$ is an identity matrix and
\begin{equation}
\label{eq:emelm_dk}
\mathbf{D}_k = \left(\left(\mathbf{I} - \HH_k\HH_{k}^{\dagger}\right)\delta\HH_k\right)^{\dagger},
\end{equation}
respectively.

Note that if $\HH_{k}^{\dagger}$ is stored, we can save some computations when adding the new nodes in the $k+1$ iteration, reducing the algorithm computational time. 

\subsection{Incremental Regularized ELM}

The I-ELM and EM-ELM algorithms are capable of increasing the hidden node number of an SLFN trained using the original ELM algorithm. However, these methods have a problem in some applications where the initial hidden layer output matrix is rank deficient, and the accuracy of its computations may be compromised \citep{Xu2016}. Also, they inherit ELM problems and sometimes cannot achieve the expected testing accuracy. 

The overfitting problem of ELM can be solved by using the structural risk minimization principle, which was used in the R-ELM method \citep{Deng2009,Martinez-Martinez2011,Huang2012}. However, the hidden node number of an SLFN trained with the original R-ELM method is a hyperparameter and cannot be increased efficiently. 

To solve this problem, the Incremental Regularized ELM (IR-ELM) was proposed \citep{Xu2016}, which is an algorithm capable of increasing the hidden node number of an SLFN trained using the R-ELM algorithm (where the considered structural risk is the $\ell_2$ norm, with $\alpha = 0$ and $\beta_0 = 0$). IR-ELM considers that by adding one node, its new hidden output matrix $\HH_{k+1}$ is given by $\HH_{k+1} = [\HH_{k+1},\vv_k]$, where $\vv_k = [h(\ab_{\Ntil+1}\cdot\xx_1 + b_{\Ntil+1}), \cdots, h(\ab_{\Ntil+}\cdot\xx_N + b_{\Ntil+1})]\transpose$ and the new output weights of the SLFN can be calculated as 
\begin{equation}
\label{eq:irelmbeta}
\bm{\beta}_{k+1} = \inv{\left(\frac{\bm{I}}{C} + \HH_{k+1}\transpose\HH_{k+1}\right)}\HH_{k+1}\transpose\tb = \mathbf{D}_{k+1}\tb.
\end{equation}

We can rewrite $\mathbf{D}_{k+1}$ as
\begin{align}
\mathbf{D}_{k+1} = \inv{\left(\frac{\bm{I}}{C} + \HH_{k+1}\transpose\HH_{k+1}\right)}\HH_{k+1}\transpose =& \inv{\left(\frac{\bm{I}}{C} + \begin{bmatrix} \nonumber
\HH\transpose_{k}\\
\vv_{k}\transpose
\end{bmatrix}[\HH_{k},\vv_{k}]\right)}\begin{bmatrix} 
\HH\transpose_{k}\\
\vv_{k}\transpose
\end{bmatrix}\\
=& \begin{bmatrix}
\HH_{k}\transpose\HH_{k} + \frac{\bm{I}}{C} & \HH_{k}\transpose\vv_k\\
\vv_k\transpose\HH_k & \vv_{k}\transpose\vv_k + C^{-1}
\end{bmatrix}^{-1}\begin{bmatrix} \nonumber
\HH\transpose_{k}\\
\vv_{k}\transpose
\end{bmatrix}.
\end{align}

According to \citet{Xu2016}, $\bm{\beta}_{k+1}$ can be updated using:
\begin{equation}
\label{eq:bkp1_irelm}
\bm{\beta}_{k+1} = \mathbf{D}_{k+1}\tb = \begin{bmatrix}
\mathbf{L}_{k+1}\\
\mathbf{M}_{k+1}
\end{bmatrix}\tb,
\end{equation}
where $\mathbf{M}_{k+1}$ and $\mathbf{L}_{k+1}$ are given by

\begin{equation}
\mathbf{M}_{k+1} = \frac{\vv_k\transpose \left(\mathbf{I} - \HH_{k}\mathbf{D}_{k}\right)}{\vv_k\transpose\left(\mathbf{I} - \HH_{k}\mathbf{D}_{k}\right)\vv_k + C^{-1}}
\end{equation}
and
\begin{equation}
\mathbf{L}_{k+1} = \mathbf{D}_{k}\left(\mathbf{I} - \vv_k\mathbf{M}_{k+1}\right),
\end{equation}
respectively.

By using Eq. \ref{eq:bkp1_irelm}, we can find the new network output weight ($\bbeta_{k+1}$) efficiently, using known information, which is usually faster than training a new and larger SLFN. According to \citet{Zhang2015}, methods that considers the error $\ell_2$ norm can suffer in the presence of outliers, which is the case of IR-ELM.

%% file: sec/method.tex
\section{Proposed Method}\label{sec:PropMeth}

In this section, we present our proposed methods. We first generalize OR-ELM to multi-target regression problems, presenting its ADMM update rules. This method is named Generalized Outlier Robust ELM (GOR-ELM). We also present an incremental version of this generalization.

\subsection{GOR-ELM}\label{subsec:GORELM}
Multi-target regression (MTR) refers to the prediction problems of multiple real-valued outputs. Such problems occur in different fields, including ecological modeling, economy, energy, data mining, and medical image \citep{ Ghosn1997,Dzeroski2000,Kocev2009, Wang2015a, Zhen2016, Spyromitros-Xioufis2016}.

A natural way to treat MTR problems is to obtain a model for each output, separately. However, one of the main MTR challenges is contemplating the output relationship, besides the non-linear input-output mapping, on modeling \citep{Zhang2010,Zhen2017}. 

In the age of big data, MTR problems are becoming more and more common in the massive amount of accumulated data. However, inherent to this immense volume of data, other problems arise.  Outliers are becoming more frequent due to different causes, such as instrumentation or human error \citep{Zhang2015,Hodge2004}. Although modeling the complex relationship between outputs and input-output on MTR problems are well studied, little has been done regarding outliers on MTR problems. 

The use of $\ell_1$ norm to achieve robustness to outliers in regression problems is a common practice in the literature \citep{Zhang2015,Xu2013,Chen2017a}. The idea of GOR-ELM is to extend the OR-ELM to MTR problems. Nevertheless, instead of treating each output separately, possible output relationships are considered on GOR-ELM through its regularization. Therefore, GOR-ELM is especially suitable for problems where output outliers may occur in a structured\footnote{Structured in the sense that all outputs of an outlier are simultaneously affected.} way.

Thus, our approach is an extension of OR-ELM where matrix norms are considered, instead of vector ones. We replace $\ell_1$ and $\ell_2$ norms by $\ell_{2,1}$ and Frobenius norms, respectively. We can also view our approach as an extension of GR-ELM (\equref{eq:GR_ELM_ADMM}), replacing the Frobenius norm of the error $\HH\BB-\TT$ by its $\ell_{2,1}$ norm. According to \cite{l21robust} and \cite{l21robust2}, the $\ell_{2,1}$ also offers robustness to outliers, as well as $\ell_1$ norm, which implies that GOR-ELM returns a robust network.

The following optimization problem is proposed
\begin{equation}
    \underset{\BB}{\text{minimize}}\quad \tau \norm{\HH \BB - \TT}_{2,1} + \lambda \alpha \norm{\BB}_{2,1} + \frac{\lambda}{2} (1-\alpha) \norm{\BB}_F^2.
    \label{eq:GOR_ELM}
\end{equation}

The optimization problem of GOR-ELM (\equref{eq:GOR_ELM}) is equivalent to the following constrained problem
\begin{equation}
 \begin{aligned}
 & \underset{\EE,\BB,\ZZ}{\text{minimize}}
 & & \tau\norm{\EE}_{2,1} + \frac{\lambda}{2} (1-\alpha)\norm{\BB}_F^2 +\lambda \alpha\norm{\ZZ}_{2,1} \\
 & \text{subject to}
 & & \EE = \HH\BB-\TT \\
 & & & \BB - \ZZ = \boldZero,
 \end{aligned}
 \label{eq:GOR_ELM_ADMM}
\end{equation}
where the objective function is separable on $\EE$, $\BB$, and $\ZZ$. Similar to GR-ELM, we use ADMM to solve the optimization problem. On each iteration of ADMM, the algorithm performs an alternated minimization of the augmented Lagrangian over $\EE$, $\BB$, and $\ZZ$.

Note that we can rewrite \equref{eq:GOR_ELM_ADMM} as
\begin{equation}
 \begin{aligned}
 & \underset{\EE,\BB,\ZZ}{\text{minimize}}
 & & \tau\norm{\EE}_{2,1} + \frac{\lambda(1-\alpha)}{2} \norm{\BB}_F^2 +\lambda \alpha \norm{\ZZ}_{2,1} \\
 & \text{subject to}
 & & \widetilde{\Ab}\EE + \widetilde{\BB}\BB+\widetilde{\CC}\ZZ + \widetilde{\DD} = \boldZero,
 \end{aligned}
 \label{eq:GOR_ELM_ADMM2}
\end{equation}
where $\tilde{\Ab} = \left[-\I_N \, , \, \boldZero_{N\times \Ntil}\right]\transpose$, $\widetilde{\BB} = \left[\HH\transpose \, , \, \I_{\Ntil}\right]\transpose$, $\widetilde{\CC} = \left[\boldZero_{\Ntil\times N} \, , \, -\I_{\Ntil}\right]\transpose$, and $\widetilde{\DD} = \left[-\TT\transpose \, , \, \boldZero_{m \times \Ntil}\right]\transpose$. According to \citet{chen2016direct}, the optimization problem established in  \equref{eq:GOR_ELM_ADMM2} can be solved using the 3-block ADMM algorithm, which does not necessarily converge. A sufficient condition to its convergence is that at least one of the following statements is true: $\tilde{A}\transpose\tilde{B} = \bm{0}$, $\tilde{A}\transpose\tilde{C} = \bm{0}$ or $\tilde{B}\transpose\tilde{C} = \bm{0}$. In our method, we can check that $\tilde{A}\transpose\tilde{C} = \bm{0}$ is true. Then, GOR-ELM algorithm converges. For more information, see \cite{chen2016direct}.

The augmented Lagrangian of the \equref{eq:GOR_ELM_ADMM2} is
\begin{align}
    L(\EE,\BB,\ZZ,\YY) =& \,\,\tau \norm{\EE}_{2,1} + \frac{\lambda(1-\alpha)}{2} \norm{\BB}_F^2 +\lambda \alpha \norm{\ZZ}_{2,1}\nonumber\\ 
    &+ \frac{\rho}{2}\norm{\widetilde{\Ab}\EE + \widetilde{\BB}\BB+\widetilde{\CC}\ZZ + \widetilde{\DD}}_F^2\nonumber\\
    &+ \innerprod{\YY,\widetilde{\Ab}\EE + \widetilde{\BB}\BB+\widetilde{\CC}\ZZ + \widetilde{\DD}},
    \label{eq:GOR_ELM_ADMM_AUGM_LAG}
\end{align}
where $\YY$ is the Lagrange multiplier, and $\rho > 0$ is a regularization parameter.

Using the scaled dual variable $\UU=(1/\rho) \YY$, we can rewrite the augmented Lagrangian as
\begin{align}
    L(\EE,\BB,\ZZ,\UU) =& \,\,\tau \norm{\EE}_{2,1} + \frac{\lambda(1-\alpha)}{2} \norm{\BB}_F^2 +\lambda \alpha \norm{\ZZ}_{2,1}  \nonumber\\
    &+ \frac{\rho}{2}\norm{\widetilde{\Ab}\EE + \widetilde{\BB}\BB+\widetilde{\CC}\ZZ + \widetilde{\DD}+\UU}_F^2 - \frac{\rho}{2}\norm{\UU}_F^2.
    \label{eq:GOR_ELM_ADMM_SCALED_DUAL_AUGM_LAG}
\end{align}

Since the Frobenius norm is separable, and by the equivalence of \equref{eq:GOR_ELM_ADMM} and \equref{eq:GOR_ELM_ADMM2}, the augmented Lagrangian (\equref{eq:GOR_ELM_ADMM_SCALED_DUAL_AUGM_LAG}) of GOR-ELM can be written as
\begin{align}
    L(\EE,\BB,\ZZ,\UU) =& \,\,\tau \norm{\EE}_{2,1} + \frac{\lambda(1-\alpha)}{2} \norm{\BB}_F^2 +\lambda\alpha\norm{\ZZ}_{2,1}  \nonumber\\
    &+ \frac{\rho}{2}\norm{\HH\BB-\TT-\EE+\UU_1}_F^2 + \frac{\rho}{2}\norm{\BB-\ZZ+\UU_2}_F^2 \nonumber \\
    &-\frac{\rho}{2}\norm{\UU_1}_F^2-\frac{\rho}{2}\norm{\UU_2}_F^2,
    \label{eq:GOR_ELM_ADMM_SCALED_DUAL_AUGM_LAG2}
\end{align}
where $\UU=\left[ \UU_1\transpose \quad \UU_2\transpose \right]\transpose$, with $\UU_1\in\R^{N\times m}$ and $\UU_2\in\R^{\Ntil\times m}$.

At iteration $k$, ADMM consists of the following update rules for 

\begin{enumerate}
    \item $\BB^{k+1}$, we have the following subproblem
        \begin{align}
            \BB^{k+1} := \argmin_{\BB} &\frac{\rho}{2} \norm{\HH \BB - \TT - \EE^k + \UU_1^k}_F^2 + \frac{\lambda(1-\alpha)}{2}\norm{\BB}_F^2 +\nonumber\\ &\frac{\rho}{2}\norm{\BB-\ZZ^k+\UU_2^k}_F^2.
            \label{eq:GOR_ELM_B_UPDATE}
        \end{align}
        Making the derivative of the \equref{eq:GOR_ELM_B_UPDATE} with respect to $\BB$ equals $\boldZero$, we have
        \begin{equation}
            \HH\transpose\left(\HH \BB - \left(\TT+\EE^k-\UU_1^k\right)\right) + \frac{\lambda(1-\alpha)}{\rho}\BB+\left(\BB-\ZZ^k+\UU_2^k\right)=\boldZero,
            \label{eq:GOR_ELM_B_GRAD_EQ_0}
        \end{equation}
        \noindent thus, the solution of \equref{eq:GOR_ELM_B_UPDATE} is
        \begin{equation}
            \BB^{k+1} = \inv{\left(\HH\transpose \HH + \eta \I\right)}\left[ \HH\transpose \left( \TT+\EE^k-\UU_1^k \right) + \left(\ZZ^k-\UU_2^k\right) \right],
            \label{eq:GOR_ELM_B}
        \end{equation}
        \noindent where $\eta = (\lambda(1-\alpha)+\rho)/\rho$.

    \item   $\ZZ^{k+1}$, we have the following subproblem
        \begin{equation}
            \ZZ^{k+1} := \argmin_{\ZZ} \frac{\lambda\alpha}{\rho} \norm{\ZZ}_{2,1} + \frac{1}{2} \norm{\BB^{k+1} - \ZZ + \UU_2^{k}}_F^2.
            \label{eq:GOR_ELM_Z_UPDATE}
        \end{equation}
        The optimization problem of \equref{eq:GOR_ELM_Z_UPDATE} is identical to the one established on the $\ZZ$ update rule of \citet{Inaba2018} method, showed in \algref{alg:grelm}. Therefore, the solution of \equref{eq:GOR_ELM_Z_UPDATE} is given by
        \begin{equation}
            \ZZ^{k+1} = \So_{\frac{\lambda\alpha}{\rho}}\left(\BB^{k+1}+\UU_2^k\right),
            \label{eq:GOR_ELM_Z_SOL}
        \end{equation}
        where $\So_\kappa(\Ab)$ is an operator that applies the block soft-thresholding operator \citep{Boyd2010a} $S_\kappa$ in each row of $\Ab$, which is defined as 
        \begin{equation}
        S_{\kappa}(\ab) = \left(1 - \frac{\kappa}{\norm{\ab}_2}\right)_+\ab,
        \end{equation} 
        with $S_{\kappa}(\bm{0}) = \bm{0}$ and $(d)_+ \equiv \max(0,d)$. Note that $S_{\kappa}(\ab)$ shrinks its argument to $\bm{0}$ if $\norm{\ab}_2 \leq \kappa$ and moves it by $\kappa$ units to the origin otherwise.

    \item $\EE^{k+1}$, we have the following subproblem
        \begin{equation}
            \EE^{k+1} := \argmin_{\EE} \frac{\tau}{\rho} \norm{\EE}_{2,1} + \frac{1}{2} \norm{\HH\BB^{k+1}-\TT-\EE+\UU_1^k}_F^2,
            \label{eq:GOR_ELM_E_UPDATE}
        \end{equation}
        which has the same structure as \equref{eq:GOR_ELM_Z_UPDATE}. Thus, we have the following solution
        \begin{equation}
            \EE^{k+1} = \So_{\frac{\tau}{\rho}}\left(\HH\BB^{k+1}-\TT+\UU_1^k\right).
            \label{eq:GOR_ELM_E_SOL}
        \end{equation}

    \item $\UU^{k+1}$, the Lagrange multiplier is updated by
        \begin{align}
            \UU_1^{k+1}&:= \UU_1^k + \left(\HH\BB^{k+1} - \TT - \EE^{k+1}\right) \label{eq:GOR_ELM_U1_UPDATE}\\
            \UU_2^{k+1}&:= \UU_2^k + \left(\BB^{k+1} - \ZZ^{k+1}\right).
            \label{eq:GOR_ELM_U_UPDATE}
        \end{align}
\end{enumerate}

\begin{figure}[htb]
	\begin{center}
		\resizebox{\linewidth}{!}{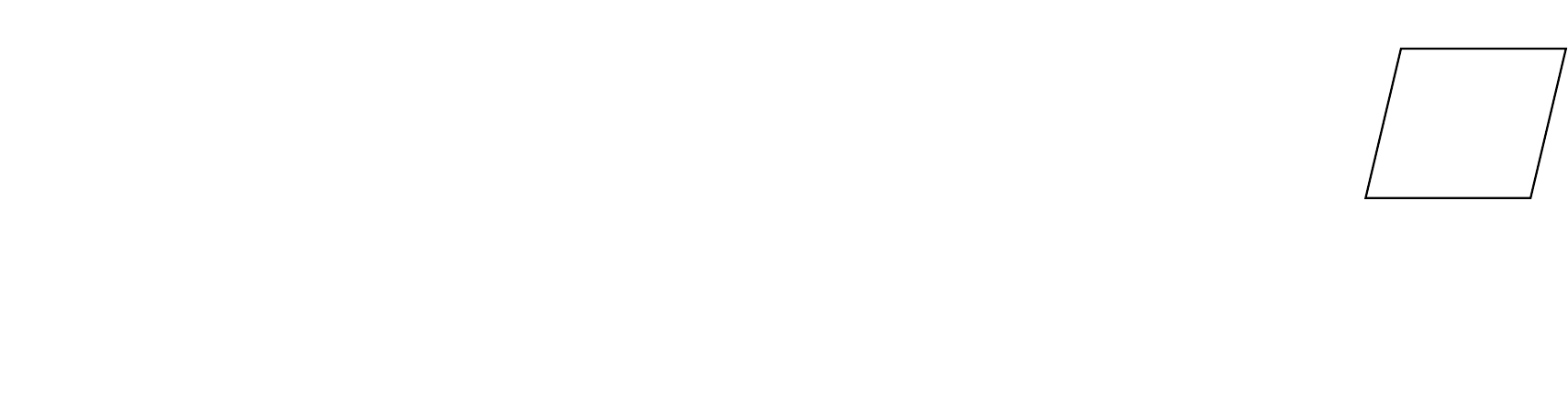}
	\end{center}
	\caption{Proposed algorithm flowchart.}
	\label{fig:flowchart}
\end{figure}

Algorithm \ref{alg:gorelm} resumes the GOR-ELM method. \figref{fig:flowchart} shows the proposed algorithm flowchart.

\begin{algorithm}[H]
	\textbf{Require:} Training samples ($\XX$,$\TT$), activation function $h(\cdot)$, number of hidden nodes $\tilde{N}$ and regularization parameters $\lambda$, $\alpha$, $\tau$ and $\rho$.\\
	\textbf{Initialization}: $\BB^0$, $\UU^0$, $\ZZ^0$, $\EE^0$ and $k=0$.\\
	Generate random weights $\mathbf{A}$ and biases $\mathbf{\nnu}$.\\
	Calculate the output of nodes $\mathbf{H}$ using \equref{eq:elm_hmatrix}.\\
	\Repeat{stopping criterion not meet}{
		$\mathbf{B}^{k+1} := \arg\min_{\mathbf{B}} \frac{\rho}{2} \norm{\mathbf{HB} - \mathbf{T} - \mathbf{E}^k + \mathbf{U}_1^k}_F^2 + \frac{\lambda(1-\alpha)}{2}\norm{\mathbf{B}}_F^2 + \frac{\rho}{2} \norm{\mathbf{B} - \mathbf{Z}^k + \mathbf{U}_2^{k}}_F^2$\\
		$\mathbf{Z}^{k+1} := \arg\min_{\mathbf{Z}} \frac{\lambda\alpha}{\rho} \norm{\textbf{Z}}_{2,1} + \frac{1}{2} \norm{\mathbf{B}^{k+1} - \mathbf{Z} + \mathbf{U}_2^{k}}_F^2$\\
		$\mathbf{E}^{k+1} := \arg\min_{\mathbf{E}} \frac{\tau}{\rho} \norm{\mathbf{E}}_{2,1} + \frac{1}{2}\norm{\mathbf{H}\mathbf{B}^{k+1} - \mathbf{T} - \mathbf{E} + \mathbf{U}_1^k}$\\
		$\mathbf{U}_1^{k+1} := \mathbf{U}_1^{k} + \rho\left(\mathbf{H}\mathbf{B}^{k+1} - \mathbf{T} - \mathbf{E}^{k+1}\right)$\\
		$\mathbf{U}_2^{k+1} := \mathbf{U}_2^{k} + \rho\left(\mathbf{B}^{k+1} - \mathbf{Z}^{k+1}\right)$\\
		k = k+1
	}
	\caption{GOR-ELM: Generalized Outlier Robust Extreme Learning Machine}
	\label{alg:gorelm}
\end{algorithm}

\subsection{Stopping Criterion}
\label{subsec:stoppcrit}
We follow the suggestions given in \cite{chen2016direct} and \cite{Boyd2009} for stopping criterion. We terminate the algorithm when primal and dual residuals, given by
	\begin{equation}
	\bm{R}^k = \tilde{\Ab}\EE^{k+1} + \tilde{\BB}\BB^{k+1} + \tilde{\CC}\ZZ^{k+1} + \tilde{\DD}
	\end{equation}
	and
	\begin{equation}
	\bm{S} = \rho\tilde{\BB}\transpose\tilde{\Ab}(\EE^{k+1} - \EE^k) + \rho\tilde{\BB}\transpose\tilde{\CC}(\ZZ^{k+1} - \ZZ^k),
	\end{equation}
	respectively, satisfy
	\begin{equation}
	\norm{\bm{R}^k}_F < \epsilon^{pri}
	\end{equation}
	and
	\begin{equation}
	\norm{\bm{S}}_F < \epsilon^{dual}.
	\end{equation}
	The tolerances $\epsilon^{pri} > 0$ and $\epsilon^{dual} > 0$ are set using the absolute and relative criterion
	\begin{equation}
	\epsilon^{pri} = \sqrt{m\Ntil}\epsilon^{abs} + \epsilon^{rel}\max(\norm{\tilde{\Ab}\EE^{k}}_F,\norm{\tilde{\BB}\BB^{k}}_F,\norm{\tilde{\CC}\ZZ^{k}}_F,\norm{\tilde{\DD}}_F),
	\end{equation}
	\begin{equation}
	\epsilon^{dual} = \sqrt{m\Ntil}\epsilon^{abs} + \epsilon^{rel}\rho\norm{\HH\transpose\UU_1 + \UU_2}_F,
	\end{equation}
	where $\epsilon^{abs} > 0$ and $\epsilon^{rel} > 0$ are the absolute and relative tolerance, respectively.

\subsection{Algorithm Complexity}
When analyzing Algorithm \ref{alg:gorelm}, we can see that the most expensive steps are the $\BB$, $\ZZ$ and $\EE$ updates, using Eqs. \ref{eq:GOR_ELM_B}, \ref{eq:GOR_ELM_Z_SOL} and \ref{eq:GOR_ELM_E_SOL}, respectively. When considering \equref{eq:GOR_ELM_B}, the most expensive operation is the matrix inversion, which has complexity $\mathcal{O}(\Ntil^3 + \Ntil^2N)$, according to \cite{akusok2015high}. We can consider this as the complexity of each iteration, since Eqs. \ref{eq:GOR_ELM_Z_SOL} and \ref{eq:GOR_ELM_E_SOL} are composed of the block soft-thresholding operation, which is a less complex operation.

%% file: figs/fluxograma.pdf_tex
\begingroup%
  \makeatletter%
  \providecommand\color[2][]{%
    \errmessage{(Inkscape) Color is used for the text in Inkscape, but the package 'color.sty' is not loaded}%
    \renewcommand\color[2][]{}%
  }%
  \providecommand\transparent[1]{%
    \errmessage{(Inkscape) Transparency is used (non-zero) for the text in Inkscape, but the package 'transparent.sty' is not loaded}%
    \renewcommand\transparent[1]{}%
  }%
  \providecommand\rotatebox[2]{#2}%
  \newcommand*\fsize{\dimexpr\f@size pt\relax}%
  \newcommand*\lineheight[1]{\fontsize{\fsize}{#1\fsize}\selectfont}%
  \ifx\svgwidth\undefined%
    \setlength{\unitlength}{491.95557602bp}%
    \ifx\svgscale\undefined%
      \relax%
    \else%
      \setlength{\unitlength}{\unitlength * \real{\svgscale}}%
    \fi%
  \else%
    \setlength{\unitlength}{\svgwidth}%
  \fi%
  \global\let\svgwidth\undefined%
  \global\let\svgscale\undefined%
  \makeatother%
  \begin{picture}(1,0.25617755)%
    \lineheight{1}%
    \setlength\tabcolsep{0pt}%
    \put(-0.20600037,-1.05781574){\color[rgb]{0,0,0}\makebox(0,0)[lt]{\begin{minipage}{0.09785112\unitlength}\centering \end{minipage}}}%
    \put(0,0){\includegraphics[width=\unitlength,page=1]{fluxograma.pdf}}%
    \put(0.89438968,0.19489771){\color[rgb]{0,0,0}\makebox(0,0)[lt]{\begin{minipage}{0.08087511\unitlength}\centering \end{minipage}}}%
    \put(0.2202435,0.21809179){\color[rgb]{0,0,0}\makebox(0,0)[lt]{\begin{minipage}{0.1422113\unitlength}\centering $\bm{B},\bm{Z}, \bm{E}, \bm{U_1}$ and $\bm{U_2}$ Initialization\end{minipage}}}%
    \put(0,0){\includegraphics[width=\unitlength,page=2]{fluxograma.pdf}}%
    \put(0.66331921,0.19354396){\color[rgb]{0,0,0}\makebox(0,0)[lt]{\begin{minipage}{0.1022012\unitlength}\centering  Criterion met?\end{minipage}}}%
    \put(0.45691443,0.24337806){\color[rgb]{0,0,0}\makebox(0,0)[lt]{\begin{minipage}{0.12840645\unitlength}\centering $\bm{B},\bm{Z},\bm{E}$ Estimation,\\ $\bm{U_1}$ and $\bm{U_2}$ Update (Eqs. \ref{eq:GOR_ELM_B}, \ref{eq:GOR_ELM_Z_SOL}, \ref{eq:GOR_ELM_E_SOL}, \ref{eq:GOR_ELM_U1_UPDATE} and \ref{eq:GOR_ELM_U_UPDATE})\end{minipage}}}%
    \put(0,0){\includegraphics[width=\unitlength,page=3]{fluxograma.pdf}}%
    \put(0.80429053,0.18612342){\color[rgb]{0,0,0}\makebox(0,0)[lt]{\begin{minipage}{0.0437277\unitlength}\centering Yes\end{minipage}}}%
    \put(0,0){\includegraphics[width=\unitlength,page=4]{fluxograma.pdf}}%
    \put(0.53373115,0.0357234){\color[rgb]{0,0,0}\makebox(0,0)[lt]{\begin{minipage}{0.0437277\unitlength}\centering No\end{minipage}}}%
    \put(0,0){\includegraphics[width=\unitlength,page=5]{fluxograma.pdf}}%
    \put(0.03579443,0.2133262){\color[rgb]{0,0,0}\makebox(0,0)[lt]{\begin{minipage}{0.12105052\unitlength}\centering Dataset and Model  Parameters\end{minipage}}}%
    \put(0,0){\includegraphics[width=\unitlength,page=6]{fluxograma.pdf}}%
    \put(0.88244881,0.20100805){\color[rgb]{0,0,0}\makebox(0,0)[lt]{\begin{minipage}{0.10561168\unitlength}\centering  Trained Model\end{minipage}}}%
  \end{picture}%
\endgroup%

%% file: sec/igor.tex
\subsection{IGOR-ELM}\label{subsec:IGORELM}

In some applications, we can choose an initial SLFN that achieves an undesired error when trained using any ELM-based algorithm. This can mean that there is still room for improvement in the model generalization. As discussed in \secref{sec:IELM}, one of the approaches that can improve the performance of an SLFN is to increase its number of hidden nodes. 

To train this larger model, we can simply use the same algorithm using a new, untrained, SLFN with more nodes. However, this procedure is not efficient if we already have a trained model, since we would discard all learned information. Thus, if the initial model efficiency is not sufficient, an approach to increase the number of hidden nodes taking advantage of the previous knowledge will be desirable.

In the literature, there are some ELM-based algorithms, such as I-ELM \citep{Huang2006}, EM-ELM \citep{GuoruiFeng2009} and IR-ELM \citep{Xu2016}, that are capable of adding more nodes to an existing SLFN that was trained using an ELM algorithm, taking advantage of known information to find the new model in less time. However, these methods usually suffer in the presence of outliers.

In this paper, we also propose the Incremental Generalized Outlier Robust ELM (IGOR-ELM). As the name suggests, it is an incremental version of GOR-ELM, where it is possible to add a new group of hidden nodes at each iteration to an existing SLFN and is robust to outliers. This algorithm solves almost the same optimization problem of GOR-ELM (\equref{eq:GOR_ELM}), taking advantage of a known model to train a larger one.

Since it uses the elastic net theory, GOR-ELM can prune hidden nodes. However, its optimization problem considers a trade-off between pruning nodes and the impact of this action in the model error. This implies that if no nodes were pruned, there is still room for improvement, and we can use IGOR-ELM to increase the model performance.

IGOR-ELM increases the network size until some stopping criterion is achieved. Every time a larger model is trained, previous knowledge is used as a starting point to the ADMM algorithm, used to solve the GOR-ELM method (Algorithm \ref{alg:gorelm}).

The starting point of each IGOR-ELM iteration is composed of increasing the dimensions of matrices with a direct relationship to the number of nodes (e.g., $\BB$, $\UU_2$, and $\ZZ$), using new values (e.g., zero-valued matrices) in these new elements. Then, random weights and biases of the new nodes are generated. The dimension of $\HH$ also increases, which implies that a larger matrix inversion must be done, which may increase the training time of each iteration.

When adding the $s$-th batch of nodes in an SLFN trained with IGOR-ELM, we can update the inverse of $\left(\HH_{s}\transpose \HH_{s} + \eta \I\right)$ using its Schur complement \citep{Petersen2007} efficiently, if $\inv{\JJ} = \inv{\left(\HH_{s-1}\transpose \HH_{s-1} + \eta \I\right)}$ is stored:
\begin{align}
\left(\HH_{s}\transpose \HH_{s} + \eta \I\right)^{-1} &= \begin{bmatrix}
\HH_{s-1}\transpose\HH_{s-1} + \eta\I & \HH_{s-1}\transpose\delta\HH_{s-1}\\
\delta\HH_{s-1}\transpose\HH_{s-1} & \delta\HH_{s-1}\transpose\delta\HH_{s-1} + \eta\I
\end{bmatrix}^{-1}\noindent \\
&= \begin{bmatrix}
\JJ & \KK\\
\KK^{\mathsf{T}} &\LL\\
\end{bmatrix}^{-1} = \begin{bmatrix}
\JJ^{'} & \KK^{'}\\
\KK^{'^{\mathsf{T}}} &\LL^{'}\\
\end{bmatrix}
\label{eq:schur}
\end{align}
where
\begin{align}
\JJ^{'} &= \inv{\JJ} + \inv{\JJ}\KK\inv{\left(\LL - \KK\transpose\inv{\JJ}\KK\right)}\KK\transpose\inv{\JJ},\\
\KK^{'} &= -\inv{\JJ}\KK\inv{\left(\LL - \KK\transpose\JJ^{-1}\KK\right)},\\
\LL^{'} &= \inv{\left(\LL - \KK\transpose\JJ^{-1}\KK\right)},
\end{align}
\noindent and $\HH_{s} = [\HH_{s-1}, \delta\HH_{s-1}]$.

Then, by adjusting variables whose dimension depends on the number of nodes and updating $\inv{\left(\HH_{s}\transpose \HH_{s} + \eta \I\right)}$, the same algorithm of GOR-ELM is used and its solution is found by using its update rules.

As the stopping criterion of the IGOR-ELM algorithm, we can chose hyperparameters such as the maximum number of total hidden nodes, an expected value for the model efficiency (e.g. a metric), or the ratio of pruned nodes. Algorithm \ref{alg:igorelm} resumes the IGOR-ELM method.\\

\begin{algorithm}[H]
	\textbf{Require:} Training samples ($\XX,\TT$), activation function $h(\cdot)$, initial number of hidden nodes $\tilde{N}$, number of new hidden nodes per iteration $\tilde{N}_{new}$ and regularization parameters $\lambda$, $\alpha$, $\tau$ and $\rho$.\\
	\textbf{Initialization}: Run Algorithm \ref{alg:gorelm} and save matrices $\mathbf{U}_2$ and $\mathbf{Z}$\\
	\While{stopping criterion not meet}{
		Generate new random weights $\mathbf{A}_{new}$ and biases $\mathbf{\nnu}_{new}$, corresponding to the new nodes.\\
		Calculate the output of new nodes $\mathbf{H}_{new}$ and update $\mathbf{H} = \left[\mathbf{H}_{old}, \mathbf{H}_{new}\right]$.\\
		Update Lagrange multiplier $\mathbf{U}_2 = \left[\mathbf{U}_2\transpose, \mathbf{0}\transpose\right]\transpose$ and $\mathbf{Z} = \left[\mathbf{Z}\transpose, \mathbf{0}\transpose\right]\transpose$, where $\mathbf{0} \in \mathds{R}^{\tilde{N}_{new}\times m}$ is a zero-valued matrix.\\
		Update the inverse of $\left(\HH_{s}\transpose \HH_{s} + \eta \I\right)$ using Eq. \ref{eq:schur}.\\
		\Repeat{GOR-ELM stopping criterion not meet}{
			$\mathbf{B}^{k+1} := \arg\min_{\mathbf{B}} \frac{\rho}{2} \norm{\mathbf{HB} - \mathbf{T} - \mathbf{E}^k + \mathbf{U}_1^k}_F^2 + \frac{\lambda(1-\alpha)}{2}\norm{\mathbf{B}}_F^2 + \frac{\rho}{2} \norm{\mathbf{B} - \mathbf{Z}^k + \mathbf{U}_2^{k}}_F^2$\\
			$\mathbf{Z}^{k+1} := \arg\min_{\mathbf{Z}} \frac{\lambda\alpha}{\rho} \norm{\textbf{Z}}_{2,1} + \frac{1}{2} \norm{\mathbf{B}^{k+1} - \mathbf{Z} + \mathbf{U}_2^{k}}_F^2$\\
			$\mathbf{E}^{k+1} := \arg\min_{\mathbf{E}} \frac{\tau}{\rho} \norm{\mathbf{E}}_{2,1} + \frac{1}{2}\norm{\mathbf{H}\mathbf{B}^{k+1} - \mathbf{T} - \mathbf{E} + \mathbf{U}_1^k}$\\
			$\mathbf{U}_1^{k+1} := \mathbf{U}_1^{k} + \rho\left(\mathbf{H}\mathbf{B}^{k+1} - \mathbf{T} - \mathbf{E}^{k+1}\right)$\\
			$\mathbf{U}_2^{k+1} := \mathbf{U}_2^{k} + \rho\left(\mathbf{B}^{k+1} - \mathbf{Z}^{k+1}\right)$\\
			k = k+1
		}
	}
	\caption{IGOR-ELM: Incremental Generalized Outlier Robust Extreme Learning Machine}
	\label{alg:igorelm}
\end{algorithm}

%% file: sec/experiments.tex
\section{Experiments}\label{sec:Exper}

To evaluate the proposed GOR-ELM and IGOR-ELM methods, we selected 15 public real-word datasets (\tabref{tab:datasets}) for multi-target regression (MTR) from the Mulan Java library \citep{Tsoumakas2010}, excluding the ones with missing data. The datasets are randomly divided into two parts: a training subset with 2/3 of the samples and a testing one with the remaining. 

\begin{table}[htb]
	\centering
	\caption{Information about multi-target regression datasets.}
	\begin{tabular}{ccccc}
		\toprule
		Dataset & \#Training Data & \#Test Data & \#Attributes & \#Targets \\ \hline
		andro   & 33              & 16          & 30           & 6         \\
		atp1d   & 225             & 112         & 411          & 6         \\
		atp7d   & 197             & 99          & 411          & 6         \\
		edm     & 103             & 51          & 16           & 2         \\
		enb     & 512             & 256         & 8            & 2         \\
		jura    & 239             & 120         & 15           & 3         \\
		oes10   & 269             & 134         & 298          & 16        \\
		oes97   & 223             & 111         & 263          & 16        \\
		rf1     & 6003            & 3002        & 64           & 8         \\
		rf2     & 5119            & 2560        & 576          & 8         \\
		scm1d   & 6535            & 3268        & 280          & 16        \\
		scm20d  & 5977            & 2989        & 61           & 16        \\
		scpf    & 95              & 48          & 23           & 3         \\
		slump   & 69              & 34          & 7            & 3         \\
		wq      & 707             & 353         & 16           & 4         \\ 
		\bottomrule
	\end{tabular}
	\label{tab:datasets}
\end{table}

We consider that a sample is an outlier based on boxplot analysis, as defined by \citet{tukey77}. Considering $Q_1$ and $Q_3$ as the first and third quartile, and $IQR = Q_3 - Q_1$ as the interquartile range of the boxplot of an attribute, a sample is considered an outlier if it is located in one of the following two intervals: $[Q_1 - 3\cdot IQR,Q_1 - 1.5\cdot IQR]$ and $[Q_3 + 1.5\cdot IQR,Q_3 + 3\cdot IQR]$. 

To evaluate the outlier robustness of our methods, we contaminated the MTR datasets using the following procedure: a subset of samples of the training set is randomly selected, which size is defined according to the outlier ratio; then each attribute of the corresponding targets of such subset are replaced by random values from $\mathcal{U}(Q_1 - 3\cdot IQR,Q_1 - 1.5\cdot IQR)$ or $\mathcal{U}(Q_3 + 1.5\cdot IQR,Q_3 + 3\cdot IQR)$. We tested our algorithms using outlier ratios of 20\% and 40\%, along with the uncontaminated datasets. This procedure was only used in the training subset.

The inputs and targets were normalized to have values in the interval $[-1,1]$ using each respective minimum and maximum values on the training dataset. This normalization was also used in the testing dataset, using the same factors. Since R-ELM, OR-ELM and IR-ELM methods were defined to one-dimensional outputs and we used multi-target datasets, they were adapted to deal with each dimension separately, where each $\bbeta_i$ is found considering only the $i$-th output, i.e,

\begin{equation}
\centering
\hat{\bbeta}_i = \inv{\left(\HH\transpose\HH + \frac{\I}{C}\right)}\HH\transpose\tb_i,
\end{equation}

\noindent and the output matrix $\hat{\BB}$ is constructed:

\begin{equation}
\centering
\hat{\BB} = [\hat{\bbeta}_1,\ldots,\hat{\bbeta}_m].
\end{equation}

Our tests were conducted in an Ubuntu 18.04 Linux computer with Intel\textregistered\ Core\texttrademark\ i7-8700K CPU with 32 GB of RAM, using MATLAB 9.4.

\subsection{Parameter Specification}
\label{subsec:param}
The number of neurons $\Ntil$ in a SLFN hidden layer defines the model complexity and has a significant impact on the underfitting/overfitting trade-off, unless some regularization is used. In ELM optimization, regularization parameters such as $\tau$, $C$ and $\lambda$ controls the trade-off between model complexity and training error when solving the associated optimization problem. In GR-ELM and GOR-ELM, the parameter $\alpha \in [0,1]$ controls the sparsity of the network: choosing $\alpha=1$ means that we prefer a sparse network over a dense network. These parameters are usually chosen using grid-search and a cross-validation method, such as $k$-fold, as done by \cite{Inaba2018}, \cite{Zhang2015}, \cite{Deng2009}, \cite{Huang2004} and others.

Note that some regularization parameters can have fixed values (e.g. 1) and the resulting minimization problem is equivalent, i.e., the result does not change if the problem is divided by a constant. According to \cite{Inaba2018}, the value of $\rho$ in ADMM algorithms is usually fixed in $\rho=1$.

First, we followed some decisions of \cite{Zhang2015}: in OR-ELM, we used 20 iterations as its stopping criterion, fixed $\tau=1$ and chose $C$ by using $5$-fold and the grid $\mathcal{G} = \{2^{-20},2^{-19},\ldots,2^{20}\}\footnote{We used a symmetric version of \cite{Zhang2015} grid.}$.

We also followed some decisions of \cite{Inaba2018}: we chose an initial $\Ntil = 1000$ for all training algorithms and datasets. When considering GR-ELM, we used primal and dual residuals as its stopping criterion, as suggested by \cite{Inaba2018}, with $\epsilon^{abs} = 10^{-3}$ and $\epsilon^{rel} = 10^{-2}$, we fixed $\rho=1$, $\lambda=1$ and obtained $C \in \mathcal{G}$ and $\alpha \in \{0,0.25,0.5,0.75,1\}$ by using $5$-fold cross-validation. We also chose R-ELM parameter $C \in \mathcal{G}$ by using the same method, and fixed $\alpha = 0$ and $\beta_0 = 0$.

When considering GOR-ELM, we fixed $\tau=1$, $\rho=1$ and obtained $\alpha \in \{0,0.25,0.5,0.75,1\}$ and $\lambda \in \mathcal{G}$ by using cross-validation. We used a combination of the stopping criterion proposed in \secref{subsec:stoppcrit} with the same values of $\epsilon^{abs}$ and $\epsilon^{rel}$ used by \cite{Inaba2018}, and a large maximum number of iterations ($k_{max} = 1000$), i.e., the algorithm stops when one of the criteria is met.

\subsection{Regression with outliers}
\label{sec:regout}


In this paper, we consider experiments with the objective of reducing the impact of initialization parameters (random weights), since our methods are non-deterministic. An experiment is defined as the following procedure: The weights between input and hidden layers ($\Ab$) and the random biases ($\nnu$) are randomly generated according to a uniform distribution in the interval $[-1,1]$; The training subset is randomly permuted; A sigmoidal activation function in the hidden layer is used: $h(x) = 1/(1+\exp(-x))$; SLFNs are trained using the algorithms and parameters obtained in $5$-fold, and we obtain the average relative root mean square error (aRRMSE) metric\footnote{Since aRRMSE measures error, a smaller value means better prediction.} and time for the training and testing subsets. 

Considering $\tb^{(l)} = \left(t_1^{(l)},\ldots,t_N^{(l)}\right)\transpose$, with $l = 1,2,\ldots,m$, as the $l$-th of the $m$ targets of the problem, aRRMSE is defined as the mean of each target relative root mean square error (RRMSE):
\begin{align}
aRRMSE(\hat{\TT},\TT) &= \frac1m\sum_{l=1}^{m} RRMSE(\hat{\tb}^{(l)},\tb^{(l)}) \nonumber\\
&= \frac1m\sum_{l=1}^{m} \sqrt{\frac{\sum_{i=1}^{N} (\hat{t_i}^{(l)} - t_i^{(l)})^2}{
\sum_{i=1}^{N} (\bar{\tb}^{(l)} - t_i^{(l)})^2
} },
\end{align}
where $\hat{\tb_i} = \left(\hat{t}_1^{(l)},\ldots,\hat{t}_N^{(l)}\right)\transpose$ is the output of a SLFN with respect to $\xx_i$, and $\bar{\tb}^{(l)} = (1/N) \sum_{i=1}^{N} t_i^{(l)}$.

\begin{figure}[htb]
	\begin{center}
		\subfloat[No Outliers]{\resizebox{0.45\linewidth}{!}{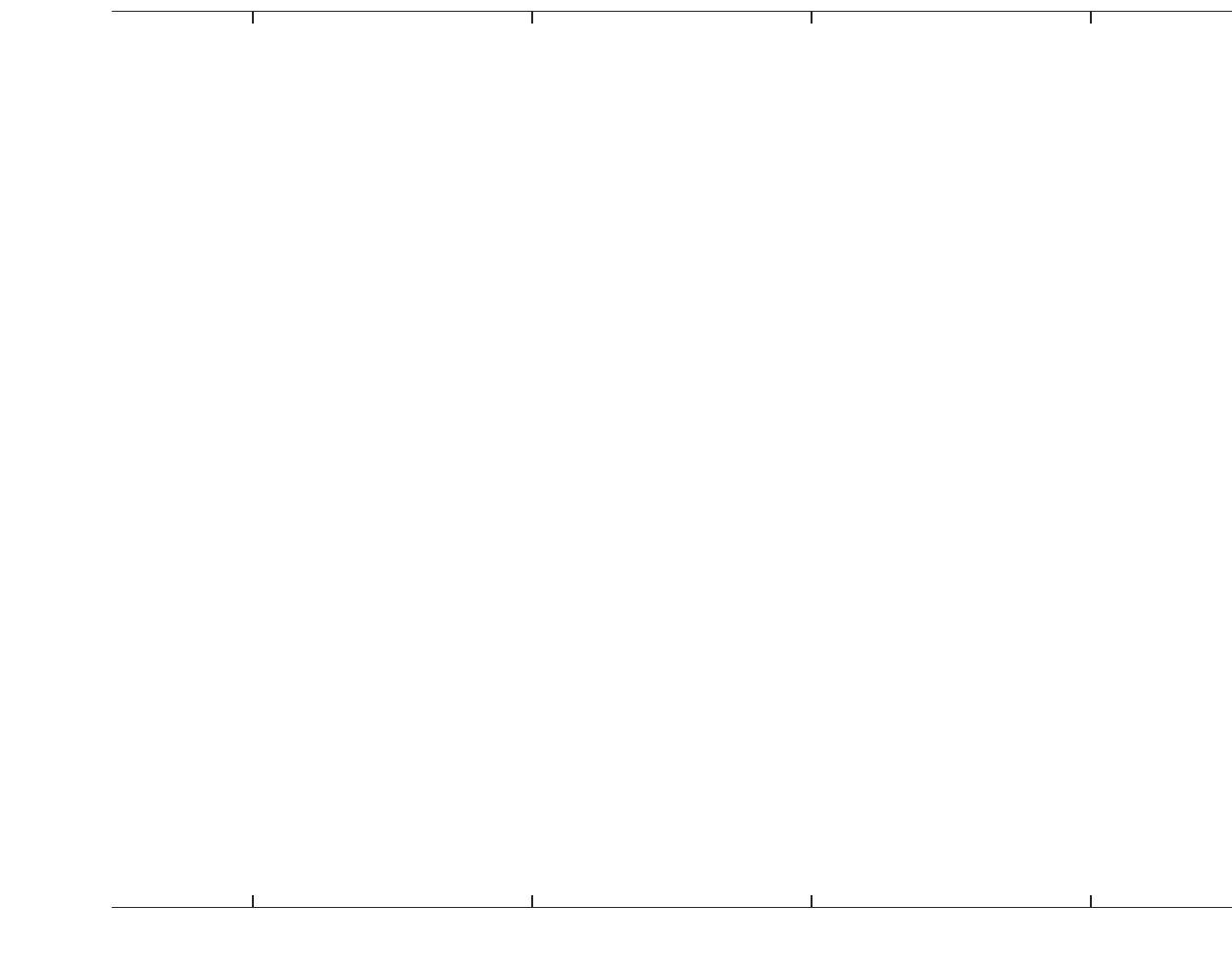}\label{fig:boxplotsa}}~
		\subfloat[20\% Outliers]{\resizebox{0.45\linewidth}{!}{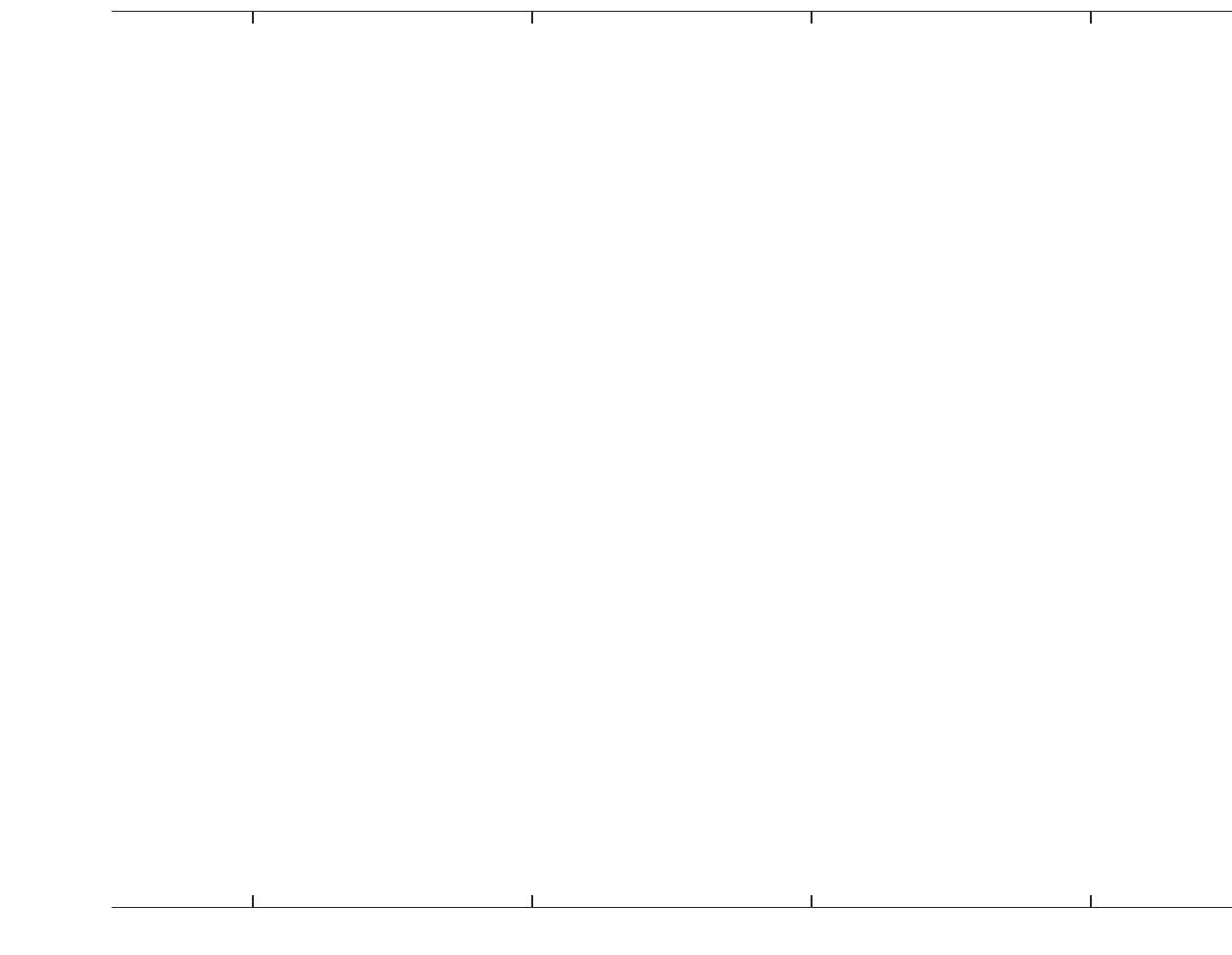}\label{fig:boxplotsb}}\\
		\subfloat[40\% Outliers]{\resizebox{0.45\linewidth}{!}{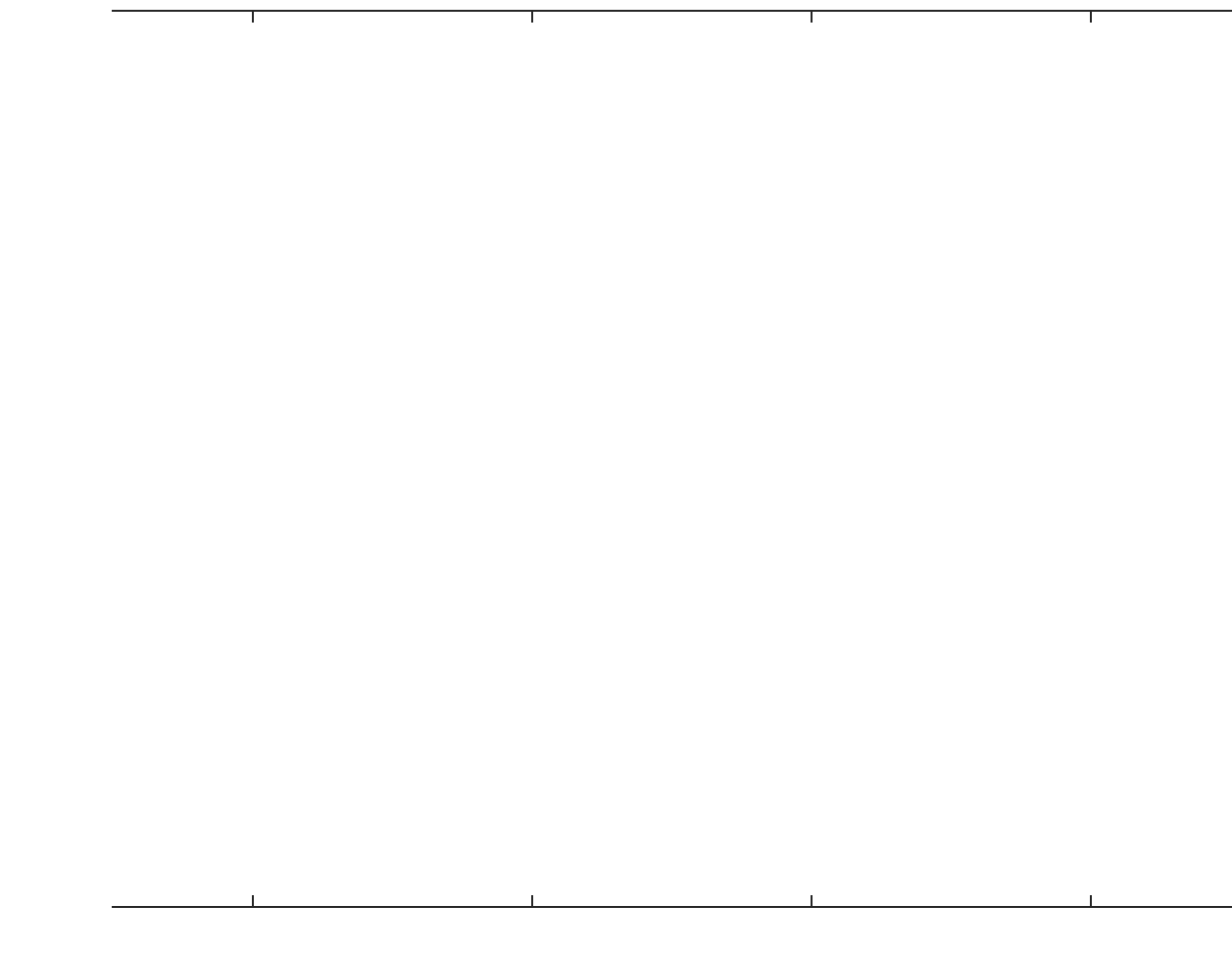}\label{fig:boxplotsc}}~
	\end{center}
	\caption{Boxplots of \tabref{tab:MTR_aRRMSE_Outlier00} (a), \tabref{tab:MTR_aRRMSE_Outlier02} (b) and \tabref{tab:MTR_aRRMSE_Outlier04} (c). The outliers were not shown for better visualization.}
	\label{fig:boxplots}
\end{figure}

The experiments were run 100 times, using the parameters discussed in \secref{subsec:param}. The average (and standard deviation) values of aRRMSE obtained in these experiments are shown in \tabref{tab:MTR_aRRMSE_Outlier00}, \ref{tab:MTR_aRRMSE_Outlier02} and \ref{tab:MTR_aRRMSE_Outlier04}, for the uncontaminated datasets and with outlier ratios of 20\% and 40\%, respectively. The best results were highlighted in boldface. We also present a boxplot of the obtained results in \figref{fig:boxplots}, where points considered outliers were hidden for better visualization.

In all datasets, similar results were obtained with all training methods, when considering the uncontaminated dataset. In this case, as shown in \tabref{tab:MTR_aRRMSE_Outlier00}, GOR-ELM achieved better metrics than the other algorithms in some datasets. Since the GOR-ELM main objective is to turn an SLFN robust to outliers, these results might imply that the datasets have some noise in its samples.

When considering that 20\% of each training subset is contaminated with outliers, R-ELM and GR-ELM achieves worse results than the robust methods. This is expected, since they use the Frobenius (or $\ell_2$) norm, which are not suitable to deal with outliers. The better results, in this case, were obtained by robust techniques, with GOR-ELM winning in 12 of the 15 datasets. When considering 40\% of outliers, similar results were obtained, with GOR-ELM achieving better aRRMSE in 9 of 15 datasets. 

It can be noted that in the aRRMSE values of GOR-ELM in the test sets remained similar in some datasets in the presence of outliers, when compared to the aRRMSE values obtained by the method when trained using the uncontaminated datasets. These results support GOR-ELM robustness to outliers characteristic.

In the presence of outliers, OR-ELM and GOR-ELM achieved better results, which was expected. Since its model considers relations between different targets (it uses matrix norms), GOR-ELM achieved better results than OR-ELM in most cases, showing that it can be a proper and robust technique in MTR tasks with the outliers inserted, according to the described methodology.

\tabref{tab:MTR_Tempo} show the time, in seconds, spent to train SLFNs with the respective algorithms. Our simulations show that GOR-ELM training stage was slightly slower than OR-ELM and GR-ELM in most cases, which was expected, since more optimization steps are needed in each iteration. Since all networks have similar architecture size, the time spent to test a set of data are almost the same. 

\tabref{tab:MTR_params} resumes the obtained parameters in the $5$-fold cross validation for the tested methods. \tabref{tab:MTR_numnodes} shows the mean number of nodes after the training stage, and we can observe that the algorithms were not capable of pruning hidden nodes (or pruned a small number of nodes) with no impact on the model performance. This also suggest that better results may be obtained by increasing the number of neurons.


\begin{table}[!htb]
	\centering
	\caption{Average aRRMSE with corresponding standard deviation for training and testing in MTR problems without outliers. Entries with $\epsilon$ indicates a value less than $10^{-3}$.}
	\resizebox{\linewidth}{!}{\begin{tabular}{ccccccccc}
			\toprule
			\multirow{2}[3]{*}{Dataset}   & \multicolumn{2}{c}{R-ELM} & \multicolumn{2}{c}{OR-ELM} & \multicolumn{2}{c}{GR-ELM}  & \multicolumn{2}{c}{GOR-ELM}\\
			\cmidrule(lr){2-3} \cmidrule(lr){4-5} \cmidrule(lr){6-7} \cmidrule(lr){8-9}
			& Train & Test & Train & Test & Train & Test & Train & Test \\
			\midrule
andro  & $\epsilon \pm \epsilon$ & $0.716 \pm 0.041$      & $0.071 \pm 0.008$ & $\bm{0.674 \pm 0.040}$ & $\epsilon \pm \epsilon$ & $0.716 \pm 0.041$      & $0.007 \pm 0.003$ & $0.718 \pm 0.042$      \\ 
atp1d  & $0.059 \pm 0.002$ & $0.404 \pm 0.014$      & $0.180 \pm 0.007$ & $\bm{0.382 \pm 0.012}$ & $0.061 \pm 0.002$ & $0.404 \pm 0.014$      & $0.170 \pm 0.007$ & $0.388 \pm 0.012$      \\ 
atp7d  & $0.091 \pm 0.002$ & $\bm{0.630 \pm 0.014}$ & $0.085 \pm 0.009$ & $0.642 \pm 0.017$      & $0.093 \pm 0.003$ & $0.630 \pm 0.014$      & $0.031 \pm 0.007$ & $0.668 \pm 0.022$      \\ 
edm    & $0.072 \pm 0.002$ & $\bm{1.404 \pm 0.080}$ & $0.082 \pm 0.004$ & $1.440 \pm 0.083$      & $0.089 \pm 0.023$ & $1.501 \pm 0.099$      & $0.076 \pm 0.004$ & $1.470 \pm 0.092$      \\ 
enb    & $0.053 \pm 0.001$ & $0.120 \pm 0.004$      & $0.061 \pm 0.002$ & $\bm{0.119 \pm 0.004}$ & $0.053 \pm 0.003$ & $0.121 \pm 0.004$      & $0.094 \pm 0.004$ & $0.140 \pm 0.006$      \\ 
jura   & $0.383 \pm 0.002$ & $0.646 \pm 0.006$      & $0.427 \pm 0.003$ & $0.633 \pm 0.006$      & $0.383 \pm 0.002$ & $0.646 \pm 0.005$      & $0.444 \pm 0.005$ & $\bm{0.628 \pm 0.008}$ \\ 
oes10  & $0.124 \pm 0.002$ & $0.361 \pm 0.006$      & $0.161 \pm 0.003$ & $0.351 \pm 0.006$      & $0.125 \pm 0.002$ & $0.360 \pm 0.006$      & $0.150 \pm 0.003$ & $\bm{0.349 \pm 0.005}$ \\ 
oes97  & $0.151 \pm 0.002$ & $0.550 \pm 0.013$      & $0.164 \pm 0.003$ & $0.562 \pm 0.014$      & $0.153 \pm 0.002$ & $\bm{0.549 \pm 0.013}$ & $0.160 \pm 0.003$ & $0.550 \pm 0.013$      \\ 
rf1    & $0.179 \pm 0.001$ & $\bm{0.219 \pm 0.031}$ & $0.195 \pm 0.002$ & $0.235 \pm 0.039$      & $0.179 \pm 0.001$ & $0.219 \pm 0.031$      & $0.180 \pm 0.002$ & $0.242 \pm 0.040$      \\ 
rf2    & $0.181 \pm 0.003$ & $0.524 \pm 0.009$      & $0.238 \pm 0.004$ & $\bm{0.193 \pm 0.010}$ & $0.181 \pm 0.003$ & $0.524 \pm 0.009$      & $0.212 \pm 0.006$ & $0.244 \pm 0.036$      \\ 
scm1d  & $0.304 \pm 0.002$ & $\bm{0.347 \pm 0.002}$ & $0.325 \pm 0.002$ & $0.352 \pm 0.002$      & $0.304 \pm 0.002$ & $0.347 \pm 0.002$      & $0.315 \pm 0.002$ & $0.347 \pm 0.002$      \\ 
scm20d & $0.398 \pm 0.002$ & $\bm{0.481 \pm 0.003}$ & $0.433 \pm 0.003$ & $0.501 \pm 0.004$      & $0.401 \pm 0.003$ & $0.482 \pm 0.003$      & $0.417 \pm 0.003$ & $0.486 \pm 0.004$      \\ 
scpf   & $0.179 \pm 0.001$ & $116.701 \pm 60.866$   & $0.246 \pm 0.002$ & $118.008 \pm 45.304$   & $0.179 \pm 0.001$ & $116.696 \pm 60.862$   & $0.281 \pm 0.002$ & $\bm{8.503 \pm 2.759}$ \\ 
slump  & $\epsilon \pm \epsilon$ & $\bm{0.892 \pm 0.029}$ & $\epsilon \pm \epsilon$ & $0.892 \pm 0.029$      & $\epsilon \pm \epsilon$ & $0.892 \pm 0.029$      & $0.011 \pm 0.006$ & $0.897 \pm 0.033$      \\ 
wq     & $0.351 \pm 0.003$ & $25.563 \pm 3.172$     & $0.404 \pm 0.004$ & $31.082 \pm 3.559$     & $0.039 \pm 0.003$ & $249.377 \pm 36.031$   & $0.720 \pm 0.001$ & $\bm{1.532 \pm 0.089}$ \\ 

			\bottomrule
	\end{tabular}}
	\label{tab:MTR_aRRMSE_Outlier00}
\end{table}

\begin{table}[!htb]
	\centering
	\caption{Average aRRMSE with corresponding standard deviation for training and testing in MTR problems with 20\% of outliers. Entries with $\epsilon$ indicates a value less than $10^{-3}$.}
	\resizebox{\linewidth}{!}{\begin{tabular}{ccccccccc}
			\toprule
			\multirow{2}[3]{*}{Dataset}   & \multicolumn{2}{c}{R-ELM} & \multicolumn{2}{c}{OR-ELM} & \multicolumn{2}{c}{GR-ELM}  & \multicolumn{2}{c}{GOR-ELM}\\
			\cmidrule(lr){2-3} \cmidrule(lr){4-5} \cmidrule(lr){6-7} \cmidrule(lr){8-9}
		& Train & Test & Train & Test & Train & Test & Train & Test \\
			\midrule
andro  & $0.000 \pm 0.000$ & $3.220 \pm 0.131$      & $0.239 \pm 0.017$ & $\bm{2.861 \pm 0.100}$ & $0.000 \pm 0.000$ & $3.220 \pm 0.131$     & $0.097 \pm 0.017$ & $3.040 \pm 0.112$      \\
atp1d  & $0.210 \pm 0.008$ & $0.887 \pm 0.058$      & $0.539 \pm 0.007$ & $0.504 \pm 0.021$      & $0.212 \pm 0.008$ & $0.884 \pm 0.058$     & $0.558 \pm 0.007$ & $\bm{0.486 \pm 0.019}$ \\
atp7d  & $0.236 \pm 0.007$ & $1.074 \pm 0.047$      & $0.361 \pm 0.015$ & $\bm{0.941 \pm 0.044}$ & $0.239 \pm 0.007$ & $1.072 \pm 0.047$     & $0.025 \pm 0.008$ & $1.474 \pm 0.086$      \\
edm    & $0.243 \pm 0.002$ & $\bm{1.307 \pm 0.064}$ & $0.285 \pm 0.004$ & $1.338 \pm 0.057$      & $0.241 \pm 0.011$ & $1.393 \pm 0.079$     & $0.278 \pm 0.005$ & $1.370 \pm 0.066$      \\
enb    & $0.613 \pm 0.003$ & $2.093 \pm 0.064$      & $0.813 \pm 0.004$ & $1.159 \pm 0.059$      & $0.603 \pm 0.003$ & $2.176 \pm 0.070$     & $0.886 \pm 0.004$ & $\bm{0.715 \pm 0.044}$ \\
jura   & $0.734 \pm 0.002$ & $0.915 \pm 0.019$      & $0.807 \pm 0.003$ & $0.746 \pm 0.013$      & $0.735 \pm 0.002$ & $0.915 \pm 0.019$     & $0.837 \pm 0.004$ & $\bm{0.679 \pm 0.013}$ \\
oes10  & $0.394 \pm 0.007$ & $0.616 \pm 0.041$      & $0.512 \pm 0.006$ & $0.463 \pm 0.021$      & $0.396 \pm 0.007$ & $0.613 \pm 0.041$     & $0.548 \pm 0.005$ & $\bm{0.399 \pm 0.011}$ \\
oes97  & $0.388 \pm 0.008$ & $0.749 \pm 0.040$      & $0.474 \pm 0.007$ & $0.641 \pm 0.022$      & $0.391 \pm 0.008$ & $0.744 \pm 0.039$     & $0.503 \pm 0.007$ & $\bm{0.586 \pm 0.015}$ \\
rf1    & $0.819 \pm 0.001$ & $0.558 \pm 0.086$      & $0.863 \pm 0.000$ & $0.242 \pm 0.038$      & $0.819 \pm 0.001$ & $0.558 \pm 0.086$     & $0.862 \pm 0.000$ & $\bm{0.238 \pm 0.040}$ \\
rf2    & $0.809 \pm 0.002$ & $0.950 \pm 0.025$      & $0.886 \pm 0.001$ & $0.285 \pm 0.011$      & $0.809 \pm 0.002$ & $0.950 \pm 0.025$     & $0.880 \pm 0.003$ & $\bm{0.282 \pm 0.021}$ \\
scm1d  & $0.878 \pm 0.002$ & $0.672 \pm 0.015$      & $0.923 \pm 0.001$ & $0.367 \pm 0.003$      & $0.878 \pm 0.002$ & $0.671 \pm 0.015$     & $0.922 \pm 0.001$ & $\bm{0.361 \pm 0.003}$ \\
scm20d & $0.856 \pm 0.003$ & $0.925 \pm 0.015$      & $0.917 \pm 0.002$ & $0.547 \pm 0.005$      & $0.857 \pm 0.003$ & $0.928 \pm 0.016$     & $0.914 \pm 0.002$ & $\bm{0.525 \pm 0.005}$ \\
scpf   & $0.145 \pm 0.002$ & $105.413 \pm 48.636$   & $0.187 \pm 0.003$ & $82.823 \pm 34.622$    & $0.145 \pm 0.002$ & $105.412 \pm 48.635$  & $0.483 \pm 0.005$ & $\bm{9.330 \pm 2.728}$ \\
slump  & $0.000 \pm 0.000$ & $3.153 \pm 0.136$      & $0.000 \pm 0.000$ & $3.153 \pm 0.137$      & $0.000 \pm 0.000$ & $3.153 \pm 0.136$     & $0.025 \pm 0.014$ & $\bm{3.138 \pm 0.155}$ \\
wq     & $0.397 \pm 0.005$ & $40.083 \pm 10.649$    & $0.433 \pm 0.007$ & $55.728 \pm 13.798$    & $0.045 \pm 0.004$ & $426.030 \pm 138.193$ & $0.800 \pm 0.002$ & $\bm{2.087 \pm 0.174}$ \\

\bottomrule
	\end{tabular}}
	\label{tab:MTR_aRRMSE_Outlier02}
\end{table}


\begin{table}[!htb]
	\centering
	\caption{Average aRRMSE with corresponding standard deviation for training and testing in MTR problems with 40\% of outliers. Entries with $\epsilon$ indicates a value less than $10^{-3}$.}
	\resizebox{\linewidth}{!}{\begin{tabular}{ccccccccc}
			\toprule
			\multirow{2}[3]{*}{Dataset}   & \multicolumn{2}{c}{R-ELM} & \multicolumn{2}{c}{OR-ELM} & \multicolumn{2}{c}{GR-ELM}  & \multicolumn{2}{c}{GOR-ELM}\\
			\cmidrule(lr){2-3} \cmidrule(lr){4-5} \cmidrule(lr){6-7} \cmidrule(lr){8-9}
			& Train & Test & Train & Test & Train & Test & Train & Test \\
			\midrule
andro  & $0.000 \pm 0.000$ & $3.295 \pm 0.215$      & $0.428 \pm 0.012$ & $\bm{2.542 \pm 0.138}$ & $0.000 \pm 0.000$ & $3.295 \pm 0.215$     & $0.267 \pm 0.016$ & $2.815 \pm 0.164$       \\
atp1d  & $0.261 \pm 0.011$ & $1.478 \pm 0.079$      & $0.666 \pm 0.013$ & $0.760 \pm 0.036$      & $0.263 \pm 0.010$ & $1.476 \pm 0.078$     & $0.709 \pm 0.012$ & $\bm{0.677 \pm 0.030}$  \\
atp7d  & $0.302 \pm 0.009$ & $1.417 \pm 0.070$      & $0.457 \pm 0.014$ & $\bm{1.237 \pm 0.061}$ & $0.305 \pm 0.009$ & $1.413 \pm 0.070$     & $0.018 \pm 0.005$ & $2.134 \pm 0.141$       \\
edm    & $0.308 \pm 0.001$ & $\bm{1.402 \pm 0.076}$ & $0.356 \pm 0.003$ & $1.444 \pm 0.081$      & $0.301 \pm 0.008$ & $1.502 \pm 0.094$     & $0.358 \pm 0.008$ & $1.479 \pm 0.086$       \\
enb    & $0.646 \pm 0.005$ & $3.187 \pm 0.090$      & $0.817 \pm 0.006$ & $2.479 \pm 0.077$      & $0.633 \pm 0.006$ & $3.297 \pm 0.097$     & $0.878 \pm 0.005$ & $\bm{1.615 \pm 0.081}$  \\
jura   & $0.792 \pm 0.003$ & $1.252 \pm 0.035$      & $0.859 \pm 0.003$ & $1.130 \pm 0.026$      & $0.793 \pm 0.003$ & $1.252 \pm 0.035$     & $0.877 \pm 0.004$ & $\bm{1.051 \pm 0.035}$  \\
oes10  & $0.604 \pm 0.011$ & $0.724 \pm 0.053$      & $0.787 \pm 0.005$ & $\bm{0.598 \pm 0.024}$ & $0.605 \pm 0.011$ & $0.723 \pm 0.053$     & $0.805 \pm 0.004$ & $0.608 \pm 0.020$       \\
oes97  & $0.515 \pm 0.013$ & $0.916 \pm 0.063$      & $0.642 \pm 0.010$ & $0.726 \pm 0.037$      & $0.518 \pm 0.013$ & $0.910 \pm 0.062$     & $0.682 \pm 0.007$ & $\bm{0.647 \pm 0.024}$  \\
rf1    & $0.920 \pm 0.001$ & $0.891 \pm 0.143$      & $0.970 \pm 0.000$ & $0.407 \pm 0.086$      & $0.920 \pm 0.001$ & $0.891 \pm 0.143$     & $0.972 \pm 0.000$ & $\bm{0.269 \pm 0.045}$  \\
rf2    & $0.875 \pm 0.002$ & $1.275 \pm 0.053$      & $0.955 \pm 0.001$ & $\bm{0.548 \pm 0.025}$ & $0.875 \pm 0.002$ & $1.275 \pm 0.053$     & $0.947 \pm 0.003$ & $0.567 \pm 0.031$       \\
scm1d  & $0.925 \pm 0.002$ & $0.899 \pm 0.020$      & $0.972 \pm 0.001$ & $0.410 \pm 0.005$      & $0.925 \pm 0.002$ & $0.898 \pm 0.020$     & $0.970 \pm 0.001$ & $\bm{0.405 \pm 0.006}$  \\
scm20d & $0.895 \pm 0.003$ & $1.262 \pm 0.023$      & $0.953 \pm 0.002$ & $\bm{0.692 \pm 0.014}$ & $0.895 \pm 0.003$ & $1.269 \pm 0.024$     & $0.945 \pm 0.002$ & $0.697 \pm 0.017$       \\
scpf   & $0.085 \pm 0.002$ & $96.842 \pm 36.668$    & $0.108 \pm 0.003$ & $104.051 \pm 44.906$   & $0.085 \pm 0.002$ & $96.825 \pm 36.658$   & $0.257 \pm 0.005$ & $\bm{18.538 \pm 7.851}$ \\
slump  & $0.001 \pm 0.000$ & $4.804 \pm 0.515$      & $0.000 \pm 0.000$ & $4.824 \pm 0.520$      & $0.001 \pm 0.000$ & $4.804 \pm 0.515$     & $0.257 \pm 0.015$ & $\bm{3.044 \pm 0.199}$  \\
wq     & $0.398 \pm 0.006$ & $66.947 \pm 25.850$    & $0.435 \pm 0.009$ & $82.457 \pm 28.984$    & $0.046 \pm 0.004$ & $522.355 \pm 222.968$ & $0.800 \pm 0.002$ & $\bm{2.315 \pm 0.333}$  \\

			\bottomrule
	\end{tabular}}
	\label{tab:MTR_aRRMSE_Outlier04}
\end{table}


\figref{fig:boxplotsa} shows the results of the tested methods when not considering outliers. In such cases, it is possible to see in the boxplots, the median of the values are almost the same for all methods (around 0.55). When considering outliers, the median becomes unbalanced, as showed in \figref{fig:boxplotsb} and \figref{fig:boxplotsc}, and robust techniques achieve smaller values than others, where GOR-ELM reaches the best one.

\begin{table}[!htb]
	\centering
	\caption{Average time (seconds) with corresponding standard deviation for training and testing in MTR problems. Smaller training times are highlighted in boldface. Entries with $\epsilon$ indicates a value less than $10^{-3}$.}
	\resizebox{\linewidth}{!}{\begin{tabular}{ccccccccc}
			\toprule
			\multirow{2}[3]{*}{Dataset}   & \multicolumn{2}{c}{R-ELM} & \multicolumn{2}{c}{OR-ELM} & \multicolumn{2}{c}{GR-ELM}  & \multicolumn{2}{c}{GOR-ELM}\\
			\cmidrule(lr){2-3} \cmidrule(lr){4-5} \cmidrule(lr){6-7} \cmidrule(lr){8-9}
			& Train & Test & Train & Test & Train & Test & Train & Test \\
			\midrule
andro  & $\epsilon \pm \epsilon$ & ${\epsilon \pm \epsilon}$ & $0.001 \pm \epsilon$ & $\epsilon \pm \epsilon$      & $0.001 \pm \epsilon$ & $\epsilon \pm \epsilon$ & $0.003 \pm 0.001$ & $\epsilon \pm \epsilon$      \\ 
atp1d  & $\bm{0.003 \pm \epsilon}$ & ${0.001 \pm \epsilon}$ & $0.008 \pm 0.001$ & $0.001 \pm \epsilon$      & $0.004 \pm \epsilon$ & $0.001 \pm \epsilon$ & $0.019 \pm 0.002$ & $0.001 \pm \epsilon$      \\ 
atp7d  & $\bm{0.002 \pm \epsilon}$ & ${0.001 \pm \epsilon}$ & $0.007 \pm 0.001$ & $0.001 \pm \epsilon$      & $0.004 \pm \epsilon$ & $0.001 \pm \epsilon$ & $0.007 \pm \epsilon$ & $0.001 \pm \epsilon$      \\ 
edm    & $\bm{0.001 \pm \epsilon}$ & ${\epsilon \pm \epsilon}$ & $0.002 \pm \epsilon$ & $\epsilon \pm \epsilon$      & $\bm{0.001 \pm \epsilon}$ & $\epsilon \pm \epsilon$ & $0.069 \pm 0.006$ & $\epsilon \pm \epsilon$      \\ 
enb    & $\bm{0.006 \pm 0.001}$ & ${0.001 \pm \epsilon}$ & $0.034 \pm 0.003$ & $0.001 \pm \epsilon$      & $0.010 \pm 0.001$ & $0.001 \pm \epsilon$ & $0.666 \pm 0.001$ & $0.001 \pm \epsilon$      \\ 
jura   & $\bm{0.002 \pm \epsilon}$ & $0.001 \pm \epsilon$      & $0.008 \pm \epsilon$ & $0.001 \pm \epsilon$      & $0.003 \pm \epsilon$ & $0.001 \pm \epsilon$ & $0.023 \pm 0.002$ & ${\epsilon \pm \epsilon}$ \\ 
oes10  & $\bm{0.003 \pm \epsilon}$ & ${0.001 \pm \epsilon}$ & $0.014 \pm 0.001$ & $0.001 \pm \epsilon$      & $0.007 \pm \epsilon$ & $0.001 \pm \epsilon$ & $0.038 \pm 0.004$ & $0.001 \pm \epsilon$      \\ 
oes97  & $\bm{0.002 \pm \epsilon}$ & ${0.001 \pm \epsilon}$ & $0.010 \pm 0.001$ & $0.001 \pm \epsilon$      & $0.005 \pm \epsilon$ & $0.001 \pm \epsilon$ & $0.024 \pm 0.003$ & $0.001 \pm \epsilon$      \\ 
rf1    & $0.159 \pm 0.012$ & $0.023 \pm 0.001$      & $0.344 \pm 0.006$ & ${0.017 \pm \epsilon}$ & $\bm{0.110 \pm 0.009}$ & $0.022 \pm 0.001$ & $4.743 \pm 0.261$ & $0.022 \pm 0.001$      \\ 
rf2    & $0.151 \pm 0.009$ & $0.023 \pm 0.001$      & $0.335 \pm 0.012$ & $0.023 \pm 0.001$      & $\bm{0.115 \pm 0.009}$ & $0.027 \pm 0.002$ & $4.411 \pm 0.324$ & ${0.020 \pm 0.003}$ \\ 
scm1d  & $0.176 \pm 0.010$ & ${0.024 \pm 0.002}$ & $0.381 \pm 0.011$ & $0.024 \pm 0.001$      & $\bm{0.133 \pm 0.012}$ & $0.024 \pm 0.002$ & $1.274 \pm 0.078$ & $0.025 \pm 0.002$      \\ 
scm20d & $0.156 \pm 0.009$ & ${0.017 \pm \epsilon}$ & $0.347 \pm 0.013$ & $0.018 \pm \epsilon$      & $\bm{0.122 \pm 0.008}$ & $0.018 \pm 0.001$ & $1.459 \pm 0.108$ & $0.018 \pm 0.001$      \\ 
scpf   & $\bm{0.001 \pm \epsilon}$ & ${\epsilon \pm \epsilon}$ & $0.002 \pm \epsilon$ & $\epsilon \pm \epsilon$      & $0.002 \pm \epsilon$ & $\epsilon \pm \epsilon$ & $0.087 \pm 0.011$ & $\epsilon \pm \epsilon$      \\ 
slump  & $\bm{0.001 \pm \epsilon}$ & ${\epsilon \pm \epsilon}$ & $0.001 \pm \epsilon$ & $\epsilon \pm \epsilon$      & $\bm{0.001 \pm \epsilon}$ & $\epsilon \pm \epsilon$ & $0.080 \pm 0.007$ & $\epsilon \pm \epsilon$      \\ 
wq     & $\bm{0.010 \pm 0.001}$ & ${0.002 \pm \epsilon}$ & $0.065 \pm 0.006$ & $0.002 \pm \epsilon$      & $2.558 \pm 0.008$ & $0.002 \pm \epsilon$ & $3.535 \pm 0.036$ & $0.002 \pm \epsilon$      \\ 

			\bottomrule
	\end{tabular}}
	\label{tab:MTR_Tempo}
\end{table}

\begin{table}[!htb]
	\centering
	\caption{Parameter specifications for MTR datasets.}
	\begin{tabular}{ccccccc}
		\toprule
		\multirow{2}{*}{Dataset} & R-ELM    & OR-ELM                      & \multicolumn{2}{c}{GR-ELM}                                                              & \multicolumn{2}{c}{GOR-ELM}   \\ \cmidrule(lr){2-2} \cmidrule(lr){3-3} \cmidrule(lr){4-5} \cmidrule(lr){6-7}
		& $C$      & \multicolumn{1}{c}{$C$} & \multicolumn{1}{c}{$C$} & \multicolumn{1}{c}{$\alpha$} & $\lambda$      & $\alpha$   \\ \cline{1-7}
andro                    & $2^{20}$ & $2^{0}$  & $2^{10}$     & $0$         & $2^{-1}$      & $0$       \\
atp1d                    & $2^{-1}$ & $2^{-5}$ & $2^{-1}$     & $0$         & $2^{5}$     & $0$          \\
atp7d                    & $2^{-2}$ & $2^{-4}$ & $2^{-2}$     & $0$         & $2^{-6}$     & $1$          \\
edm                      & $2^{9}$  & $2^{8}$  & $2^{10}$     & $0.25$       & $2^{-8}$      & $0$          \\
enb                      & $2^{16}$ & $2^{11}$ & $2^{16}$     & $0.25$       & $2^{-7}$     & $1$          \\
jura                     & $2^{1}$  & $2^{0}$  & $2^{1}$      & $0$         & $2^{-1}$     & $0.5$          \\
oes10                    & $2^{-1}$ & $2^{-5}$ & $2^{-1}$     & $0$         & $2^{5}$     & $0$          \\
oes97                    & $2^{-1}$ & $2^{-4}$ & $2^{-1}$     & $0$         & $2^{4}$     & $0$          \\
rf1                      & $2^{3}$  & $2^{2}$  & $2^{3}$      & $0$      & $2^{-3}$      & $1$          \\
rf2                      & $2^{3}$  & $2^{-1}$ & $2^{3}$      & $0$         & $2^{-1}$      & $0.75$          \\
scm1d                    & $2^{-2}$ & $2^{-4}$ & $2^{-2}$     & $0$         & $2^{3}$     & $0$          \\
scm20d                   & $2^{2}$  & $2^{1}$  & $2^{2}$      & $0.25$         & $2^{-3}$     & $1$          \\
scpf                     & $2^{19}$ & $2^{16}$ & $2^{19}$     & $0$         & $2^{-11}$     & $0.75$       \\
slump                    & $2^{20}$ & $2^{17}$ & $2^{20}$     & $0$         & $2^{-8}$     & $1$          \\
wq                       & $2^{20}$ & $2^{20}$ & $2^{20}$     & $1$         & $2^{-20}$     & $1$        \\ 
		\bottomrule
	\end{tabular}
	\label{tab:MTR_params}
\end{table}

\begin{table}[!htb]
	\centering
	\caption{Mean number of nodes of GR-ELM and GOR-ELM classifiers for MTR datasets.}
	\resizebox{\linewidth}{!}{\begin{tabular}{ccccccccc}
		\toprule
		\multirow{2}{*}{Dataset} & \multicolumn{2}{c}{No Outliers} & \multicolumn{2}{c}{20\% Outliers} & \multicolumn{2}{c}{40\% Outliers} \\
		\cmidrule(lr){2-3} \cmidrule(lr){4-5} \cmidrule(lr){6-7}
		& GR-ELM         & GOR-ELM        & GR-ELM          & GOR-ELM         & GR-ELM          & GOR-ELM         \\ \hline
andro & 1000.00 & 1000.00 & 1000.00 & 1000.00 & 1000.00 & 1000.00 \\
atp1d & 1000.00 & 1000.00 & 1000.00 & 1000.00 & 1000.00 & 1000.00 \\
atp7d & 1000.00 & 741.53 & 1000.00 & 643.58 & 1000.00 & 596.03 \\
edm & 841.12 & 1000.00 & 813.42 & 1000.00 & 854.28 & 1000.00 \\
enb & 986.58 & 559.66 & 996.24 & 481.43 & 998.34 & 550.99 \\
jura & 1000.00 & 248.06 & 1000.00 & 222.45 & 1000.00 & 262.91 \\
oes10 & 1000.00 & 1000.00 & 1000.00 & 1000.00 & 1000.00 & 1000.00 \\
oes97 & 1000.00 & 1000.00 & 1000.00 & 1000.00 & 1000.00 & 1000.00 \\
rf1 & 1000.00 & 992.63 & 1000.00 & 989.94 & 1000.00 & 970.27 \\
rf2 & 1000.00 & 706.08 & 1000.00 & 690.33 & 1000.00 & 670.40 \\
scm1d & 1000.00 & 1000.00 & 1000.00 & 1000.00 & 1000.00 & 1000.00 \\
scm20d & 999.99 & 1000.00 & 1000.00 & 1000.00 & 1000.00 & 999.96 \\
scpf & 1000.00 & 991.54 & 1000.00 & 995.69 & 1000.00 & 989.68 \\
slump & 1000.00 & 606.40 & 1000.00 & 627.44 & 1000.00 & 659.66 \\
wq & 1000.00 & 1000.00 & 1000.00 & 1000.00 & 1000.00 & 1000.00 \\
		\bottomrule
	\end{tabular}}
	\label{tab:MTR_numnodes}
\end{table}

To confirm the statistical significance of our results, we use the Friedman test \citep{Demsar2006}. We considered a null hypothesis that the four compared methods are equivalent against the alternative hypothesis that they are not, and a significance level of 10\%. When considering uncontaminated datasets, the null hypothesis was not rejected, meaning that GOR-ELM is statistically equivalent to GR-ELM, R-ELM and OR-ELM.

\begin{figure}[htb]
	\begin{center}
		\subfloat[]{\includegraphics[width=0.48\textwidth]{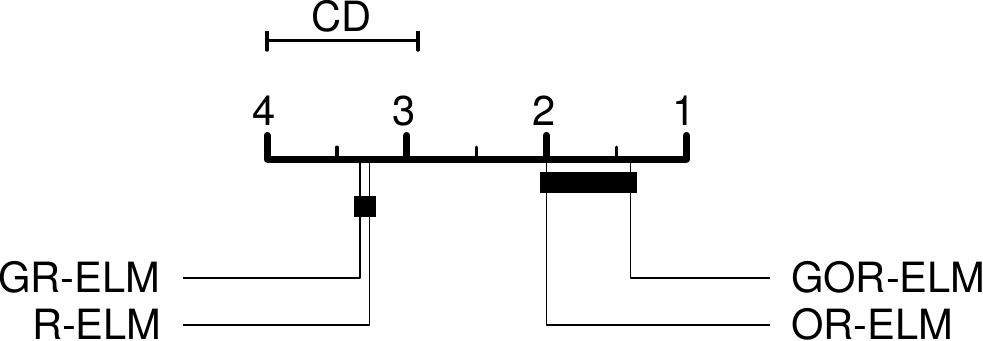}}~
		\subfloat[]{\includegraphics[width=0.48\textwidth]{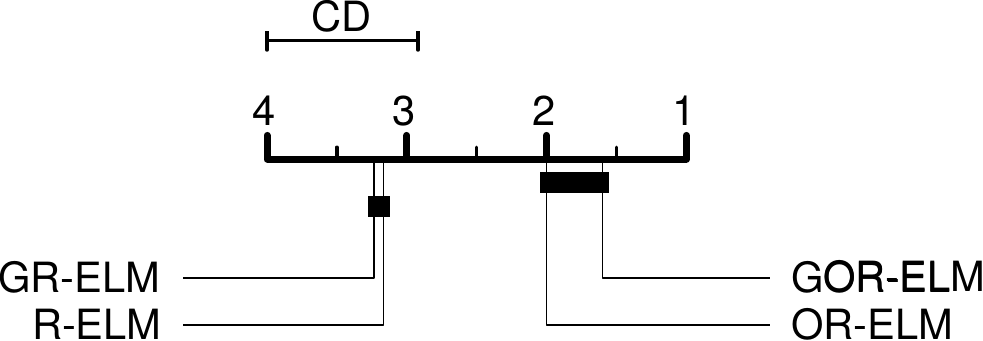}}
	\end{center}
	\caption{Critical distance considering aRRMSE in the testing subset and 20\% (a) and 40\% (b) of outliers in the training subset.}
	\label{fig:cd}
\end{figure}

However, when considering that the training dataset is contaminated with outliers, the null hypothesis is rejected with p-values of 0.000009 and 0.000271, for outlier ratio of 20\% and 40\%, respectively. Thus, we can use a post-hoc test, such as the Nemenyi test \citep{Demsar2006}. The results of this test are shown in \figref{fig:cd}, where CD means the critical distance, i.e., the minimum spacing between two methods ranks so they can be considered statistically different, and methods connected by a line are equivalent. 

The results presented in \figref{fig:cd} show that GOR-ELM is equivalent to OR-ELM and statistically better than GR-ELM and R-ELM, when the training dataset is contaminated with outliers.

\subsection{Regression with outliers - Incremental approach}

In this section, we compare IGOR-ELM only with IR-ELM in MTR tasks, since it showed better generalization performance than other incremental techniques, such as I-ELM and EM-ELM \citep{Xu2016}. We set for each method an initial number of hidden nodes ($\Ntil$) equal to 100 for all datasets and 9 batches of 100 nodes were added (up to a maximum of 1000 hidden nodes). In this test, we also specified the parameters as explained in \secref{subsec:param}.


Similar experiments to those described in \secref{sec:regout} were run 100 times, with the objective of reducing the random initialization impact of the methods, and the average (and standard deviation) values of aRRMSE obtained in these experiments are shown in \tabref{tab:MTRINC_aRRMSE_Outlier00}, \ref{tab:MTRINC_aRRMSE_Outlier02} and \ref{tab:MTRINC_aRRMSE_Outlier04}, for the original datasets and with outlier ratios of 20\% and 40\%, respectively. Similarly, the best results were highlighted in boldface.

The incremental results were similar to the ones presented in \secref{sec:regout}: When considering the uncontaminated dataset, IGOR-ELM and IR-ELM achieved close metrics, with the former reaching higher values in some cases. Considering the contaminated datasets, IGOR-ELM achieved even better results, when compared with IR-ELM. 

It can be noted that in the tested incremental approaches, a large aRRMSE variation of both techniques occurred when comparing the achieved values in uncontaminated datasets with the obtained in contaminated ones. Since this did not occur in the non-incremental approaches, it can be due to the update step (Schur Complement) of the incremental methods.

In this experiment, IGOR-ELM was the only robust method, and thus, it obtained the best values of aRRMSE in the majority of datasets, when outliers were used in the training stage. Since it uses the GOR-ELM algorithm, the outlier structures were ``detected'' and the method showed robustness to them. 

\tabref{tab:MTRINC_Tempo} show the time, in seconds, spent to train SLFNs with the respective algorithms. Our simulations show that IGOR-ELM training stage was slower than IR-ELM in most cases, which was expected, since it needs more optimization steps in each iteration.

\begin{table}[!htb]
	\centering
	\caption{Average aRRMSE with corresponding standard deviation for training and testing incremental methods in MTR problems without outliers. Entries with $\epsilon$ indicates a value less than $10^{-3}$.}
	\resizebox{\linewidth}{!}{\begin{tabular}{ccccc}
			\toprule
			\multirow{2}[3]{*}{Dataset}   & \multicolumn{2}{c}{IGOR-ELM} & \multicolumn{2}{c}{IR-ELM} \\
\cmidrule(lr){2-3} \cmidrule(lr){4-5}
			& Train & Test & Train & Test \\
			\midrule
andro & $\epsilon \pm \epsilon$ & $0.740 \pm 0.052$ & $\epsilon \pm \epsilon$ & $\bm{0.715 \pm 0.039}$ \\
atp1d & $0.170 \pm 0.009$ & $\bm{0.387 \pm 0.011}$ & $0.089 \pm 0.003$ & $0.395 \pm 0.012$ \\
atp7d & $\epsilon \pm \epsilon$ & $0.657 \pm 0.024$ & $0.058 \pm 0.002$ & $\bm{0.640 \pm 0.020}$ \\
edm & $0.086 \pm 0.004$ & $1.449 \pm 0.076$ & $0.072 \pm 0.001$ & $\bm{1.407 \pm 0.077}$ \\
enb & $0.077 \pm 0.005$ & $\bm{0.123 \pm 0.007}$ & $0.062 \pm 0.002$ & $0.123 \pm 0.004$ \\
jura & $0.449 \pm 0.002$ & $\bm{0.621 \pm 0.004}$ & $0.409 \pm 0.001$ & $0.633 \pm 0.004$ \\
oes10 & $0.150 \pm 0.003$ & $\bm{0.351 \pm 0.005}$ & $0.124 \pm 0.002$ & $0.362 \pm 0.006$ \\
oes97 & $0.160 \pm 0.003$ & $0.550 \pm 0.012$ & $0.171 \pm 0.002$ & $\bm{0.543 \pm 0.011}$ \\
rf1 & $0.179 \pm 0.002$ & $0.248 \pm 0.041$ & $0.188 \pm 0.001$ & $\bm{0.194 \pm 0.016}$ \\
rf2 & $0.212 \pm 0.006$ & $\bm{0.254 \pm 0.037}$ & $0.181 \pm 0.003$ & $0.526 \pm 0.012$ \\
scm1d & $0.315 \pm 0.001$ & $\bm{0.347 \pm 0.002}$ & $0.304 \pm 0.001$ & $0.347 \pm 0.002$ \\
scm20d & $0.417 \pm 0.003$ & $0.487 \pm 0.004$ & $0.399 \pm 0.002$ & $\bm{0.481 \pm 0.003}$ \\
scpf & $0.272 \pm 0.001$ & $\bm{38.221 \pm 16.985}$ & $0.175 \pm 0.001$ & $159.809 \pm 107.084$ \\
slump & $0.051 \pm 0.013$ & $\bm{0.881 \pm 0.061}$ & $\epsilon \pm \epsilon$ & $0.897 \pm 0.034$ \\
wq & $0.687 \pm 0.301$ & $\bm{3.501 \pm 1.460}$ & $0.397 \pm 0.042$ & $24.768 \pm 22.343$ \\
\bottomrule
	\end{tabular}}
	\label{tab:MTRINC_aRRMSE_Outlier00}
\end{table}

\begin{table}[!htb]
	\centering
	\caption{Average aRRMSE with corresponding standard deviation for training and testing of incremental methods in MTR problems with 20\% of outliers. Entries with $\epsilon$ indicates a value less than $10^{-3}$.}
	\resizebox{\linewidth}{!}{\begin{tabular}{ccccccccc}
			\toprule
			\multirow{2}[3]{*}{Dataset}   & \multicolumn{2}{c}{IGOR-ELM} & \multicolumn{2}{c}{IR-ELM} \\
\cmidrule(lr){2-3} \cmidrule(lr){4-5}
& Train & Test & Train & Test \\
\midrule
andro & $\epsilon \pm \epsilon$ & $3.256 \pm 0.171$ & $\epsilon \pm \epsilon$ & $\bm{3.220 \pm 0.135}$ \\
atp1d & $0.558 \pm 0.006$ & $\bm{0.487 \pm 0.020}$ & $0.287 \pm 0.007$ & $0.789 \pm 0.045$ \\
atp7d & $\epsilon \pm \epsilon$ & $1.430 \pm 0.102$ & $0.161 \pm 0.006$ & $\bm{1.183 \pm 0.069}$ \\
edm & $0.297 \pm 0.006$ & $1.340 \pm 0.058$ & $0.243 \pm 0.002$ & $\bm{1.296 \pm 0.063}$ \\
enb & $0.876 \pm 0.005$ & $\bm{0.792 \pm 0.054}$ & $0.638 \pm 0.004$ & $1.913 \pm 0.055$ \\
jura & $0.842 \pm 0.002$ & $\bm{0.659 \pm 0.006}$ & $0.761 \pm 0.002$ & $0.860 \pm 0.012$ \\
oes10 & $0.547 \pm 0.005$ & $\bm{0.400 \pm 0.011}$ & $0.394 \pm 0.007$ & $0.616 \pm 0.040$ \\
oes97 & $0.503 \pm 0.005$ & $\bm{0.584 \pm 0.017}$ & $0.433 \pm 0.007$ & $0.662 \pm 0.027$ \\
rf1 & $0.874 \pm 0.042$ & $\bm{0.335 \pm 0.211}$ & $0.825 \pm 0.001$ & $0.483 \pm 0.063$ \\
rf2 & $0.880 \pm 0.003$ & $\bm{0.284 \pm 0.025}$ & $0.809 \pm 0.002$ & $0.951 \pm 0.034$ \\
scm1d & $0.922 \pm 0.001$ & $\bm{0.361 \pm 0.003}$ & $0.878 \pm 0.002$ & $0.670 \pm 0.012$ \\
scm20d & $0.914 \pm 0.002$ & $\bm{0.525 \pm 0.005}$ & $0.856 \pm 0.003$ & $0.927 \pm 0.016$ \\
scpf & $0.309 \pm 0.009$ & $\bm{32.769 \pm 9.623}$ & $0.132 \pm 0.002$ & $140.212 \pm 103.234$ \\
slump & $0.097 \pm 0.030$ & $\bm{2.920 \pm 0.213}$ & $\epsilon \pm \epsilon$ & $3.167 \pm 0.807$ \\
wq & $0.729 \pm 0.004$ & $\bm{5.102 \pm 0.848}$ & $0.456 \pm 0.138$ & $41.922 \pm 81.100$ \\

			\bottomrule
	\end{tabular}}
	\label{tab:MTRINC_aRRMSE_Outlier02}
\end{table}

\begin{table}[!htb]
	\centering
	\caption{Average aRRMSE with corresponding standard deviation for training and testing of incremental methods in MTR problems with 40\% of outliers. Entries with $\epsilon$ indicates a value less than $10^{-3}$.}
	\resizebox{\linewidth}{!}{\begin{tabular}{ccccccccc}
			\toprule
			\multirow{2}[3]{*}{Dataset}   & \multicolumn{2}{c}{IGOR-ELM} & \multicolumn{2}{c}{IR-ELM} \\
\cmidrule(lr){2-3} \cmidrule(lr){4-5}
& Train & Test & Train & Test \\
\midrule
andro & $\epsilon \pm \epsilon$ & $3.329 \pm 0.231$ & $\epsilon \pm \epsilon$ & $\bm{3.298 \pm 0.199}$ \\
atp1d & $0.707 \pm 0.011$ & $\bm{0.677 \pm 0.030}$ & $0.355 \pm 0.010$ & $1.315 \pm 0.072$ \\
atp7d & $\epsilon \pm \epsilon$ & $2.112 \pm 0.157$ & $0.208 \pm 0.008$ & $\bm{1.565 \pm 0.092}$ \\
edm & $0.372 \pm 0.006$ & $1.455 \pm 0.073$ & $0.308 \pm 0.001$ & $\bm{1.408 \pm 0.071}$ \\
enb & $0.841 \pm 0.006$ & $\bm{2.202 \pm 0.078}$ & $0.676 \pm 0.005$ & $2.933 \pm 0.085$ \\
jura & $0.898 \pm 0.002$ & $\bm{0.975 \pm 0.016}$ & $0.822 \pm 0.002$ & $1.185 \pm 0.028$ \\
oes10 & $0.805 \pm 0.004$ & $\bm{0.605 \pm 0.019}$ & $0.604 \pm 0.011$ & $0.725 \pm 0.054$ \\
oes97 & $0.681 \pm 0.006$ & $\bm{0.647 \pm 0.024}$ & $0.571 \pm 0.009$ & $0.792 \pm 0.045$ \\
rf1 & $0.970 \pm \epsilon$ & $\bm{0.321 \pm 0.061}$ & $0.926 \pm 0.001$ & $0.754 \pm 0.082$ \\
rf2 & $0.947 \pm 0.003$ & $\bm{0.567 \pm 0.034}$ & $0.876 \pm 0.002$ & $1.272 \pm 0.047$ \\
scm1d & $0.970 \pm 0.001$ & $\bm{0.406 \pm 0.007}$ & $0.925 \pm 0.002$ & $0.901 \pm 0.020$ \\
scm20d & $0.944 \pm 0.003$ & $\bm{0.696 \pm 0.021}$ & $0.895 \pm 0.003$ & $1.263 \pm 0.022$ \\
scpf & $0.149 \pm 0.004$ & $\bm{62.009 \pm 25.653}$ & $0.076 \pm 0.002$ & $122.035 \pm 50.450$ \\
slump & $0.325 \pm 0.032$ & $\bm{3.039 \pm 0.313}$ & $0.001 \pm \epsilon$ & $4.986 \pm 0.559$ \\
wq & $0.735 \pm 0.013$ & $\bm{5.718 \pm 1.343}$ & $0.445 \pm 0.018$ & $53.181 \pm 27.679$ \\

			\bottomrule
	\end{tabular}}
	\label{tab:MTRINC_aRRMSE_Outlier04}
\end{table}

\begin{table}[!htb]
	\centering
	\caption{Average time (seconds) with corresponding standard deviation for training and testing of incremental methods in MTR problems. Smaller training times are highlighted in boldface. Entries with $\epsilon$ indicates a value less than $10^{-3}$.}
	\resizebox{\linewidth}{!}{\begin{tabular}{ccccccccc}
			\toprule
			\multirow{2}[3]{*}{Dataset}   & \multicolumn{2}{c}{IGOR-ELM} & \multicolumn{2}{c}{IR-ELM} \\
\cmidrule(lr){2-3} \cmidrule(lr){4-5}
& Train & Test & Train & Test \\
\midrule
andro & $0.096 \pm 0.003$ & ${\epsilon \pm \epsilon}$ & $\bm{0.009 \pm \epsilon}$ & $\epsilon \pm \epsilon$ \\
atp1d & $0.082 \pm 0.004$ & ${0.001 \pm \epsilon}$ & $\bm{0.027 \pm 0.001}$ & $0.001 \pm \epsilon$ \\
atp7d & $0.125 \pm 0.006$ & ${0.001 \pm \epsilon}$ & $\bm{0.025 \pm 0.001}$ & $0.001 \pm \epsilon$ \\
edm & $0.604 \pm 0.019$ & ${\epsilon \pm \epsilon}$ & $\bm{0.015 \pm \epsilon}$ & $\epsilon \pm \epsilon$ \\
enb & $5.676 \pm 0.101$ & ${0.001 \pm \epsilon}$ & $\bm{0.050 \pm 0.005}$ & $0.001 \pm \epsilon$ \\
jura & $0.089 \pm 0.004$ & ${0.001 \pm \epsilon}$ & $\bm{0.022 \pm \epsilon}$ & $0.001 \pm \epsilon$ \\
oes10 & $0.088 \pm 0.004$ & ${0.001 \pm \epsilon}$ & $\bm{0.027 \pm 0.002}$ & $0.001 \pm \epsilon$ \\
oes97 & $0.090 \pm 0.004$ & ${0.001 \pm \epsilon}$ & $\bm{0.024 \pm \epsilon}$ & $0.001 \pm \epsilon$ \\
rf1 & $14.164 \pm 5.781$ & $0.021 \pm 0.002$ & $\bm{4.252 \pm 0.043}$ & ${0.019 \pm 0.001}$ \\
rf2 & $8.179 \pm 0.325$ & $0.027 \pm 0.004$ & $\bm{3.202 \pm 0.048}$ & ${0.022 \pm 0.002}$ \\
scm1d & $\bm{3.473 \pm 0.152}$ & $0.029 \pm 0.003$ & $5.092 \pm 0.063$ & ${0.022 \pm 0.001}$ \\
scm20d & $6.483 \pm 0.427$ & ${0.017 \pm 0.001}$ & $\bm{4.243 \pm 0.042}$ & $0.018 \pm 0.001$ \\
scpf & $0.788 \pm 0.058$ & ${\epsilon \pm \epsilon}$ & $\bm{0.014 \pm \epsilon}$ & $\epsilon \pm \epsilon$ \\
slump & $1.206 \pm 0.119$ & ${\epsilon \pm \epsilon}$ & $\bm{0.012 \pm \epsilon}$ & $\epsilon \pm \epsilon$ \\
wq & $11.521 \pm 0.145$ & ${0.002 \pm \epsilon}$ & $\bm{0.072 \pm 0.003}$ & $0.002 \pm \epsilon$ \\

			\bottomrule
	\end{tabular}}
	\label{tab:MTRINC_Tempo}
\end{table}

\begin{figure}[htb]
	\begin{center}
		\subfloat[No Outliers]{\resizebox{0.45\linewidth}{!}{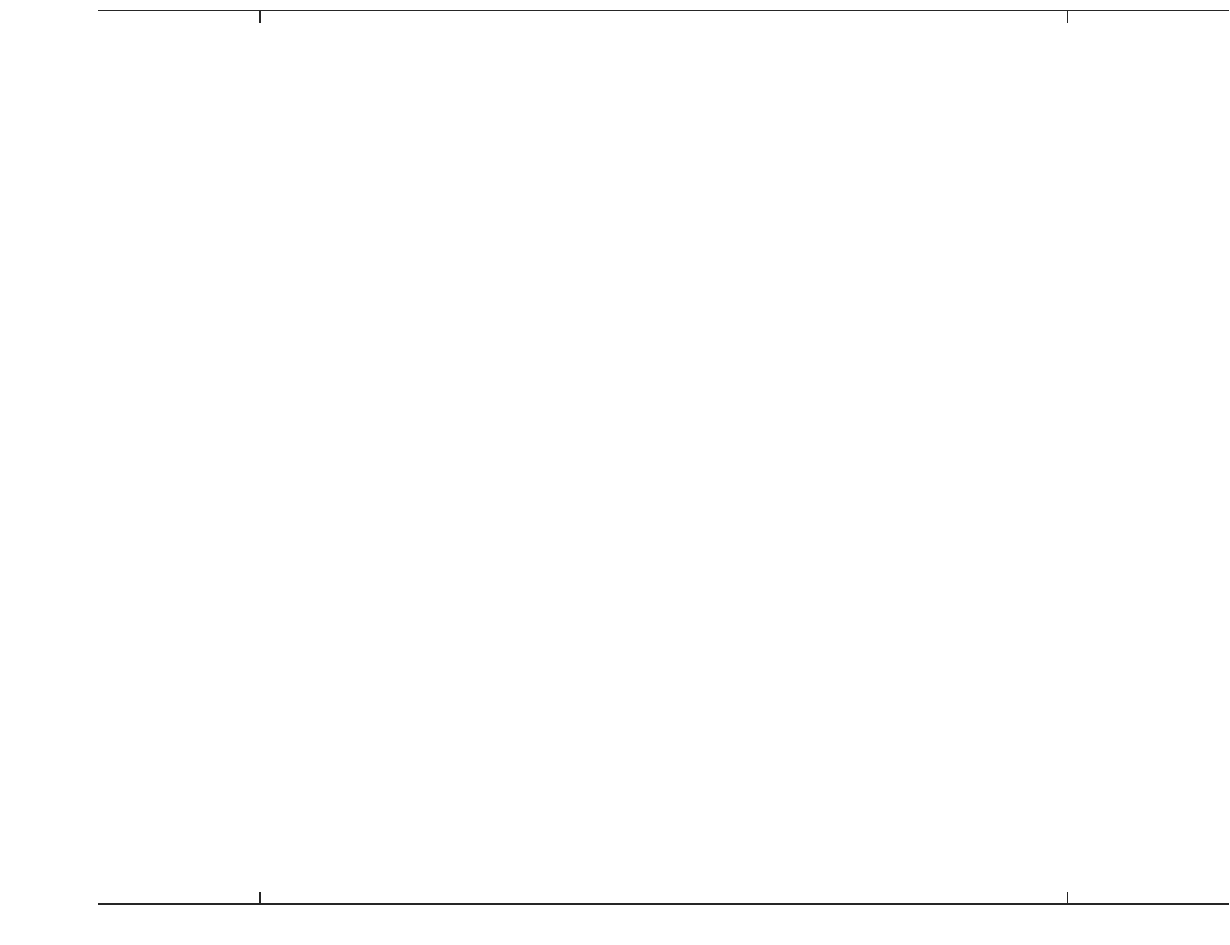}}~
		\subfloat[20\% Outliers]{\resizebox{0.45\linewidth}{!}{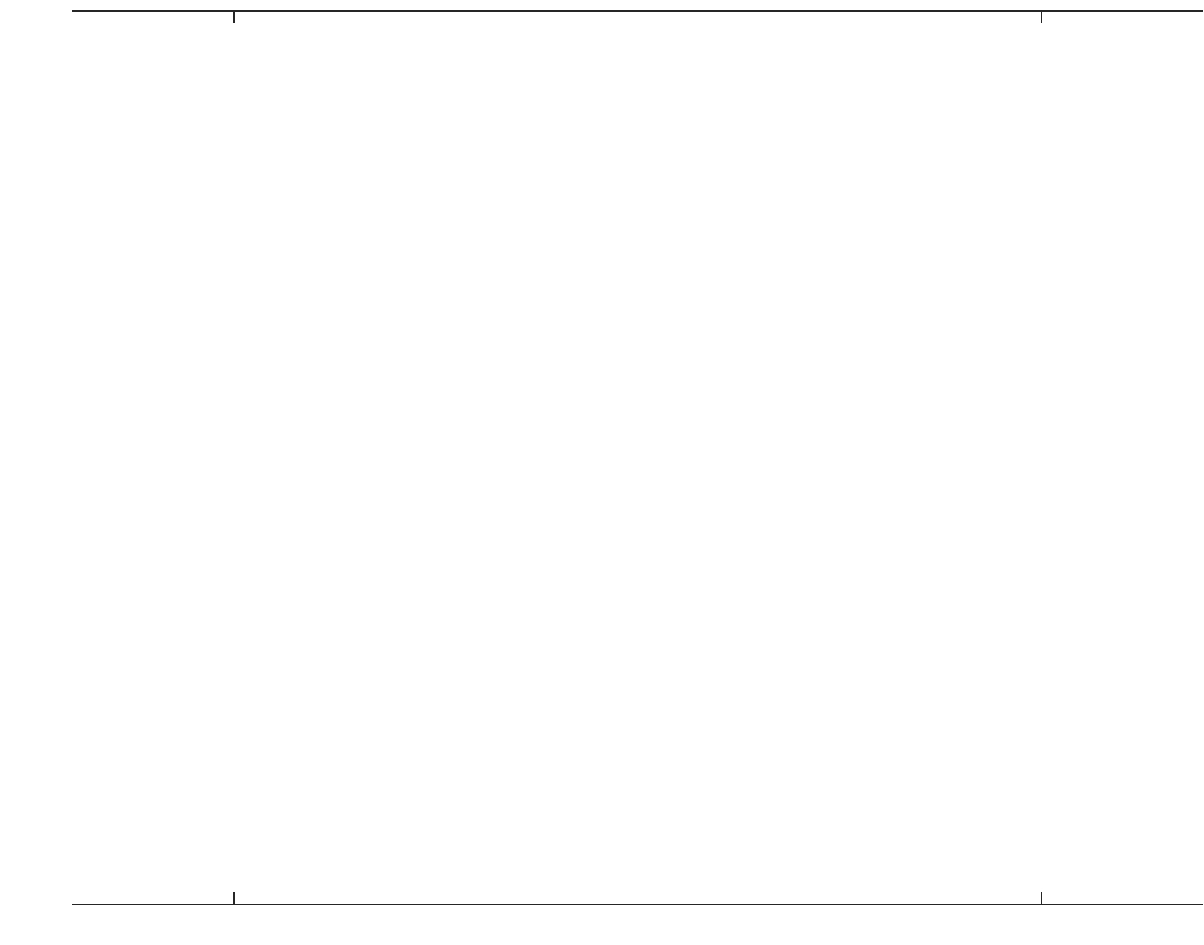}}\\
		\subfloat[40\% Outliers]{\resizebox{0.45\linewidth}{!}{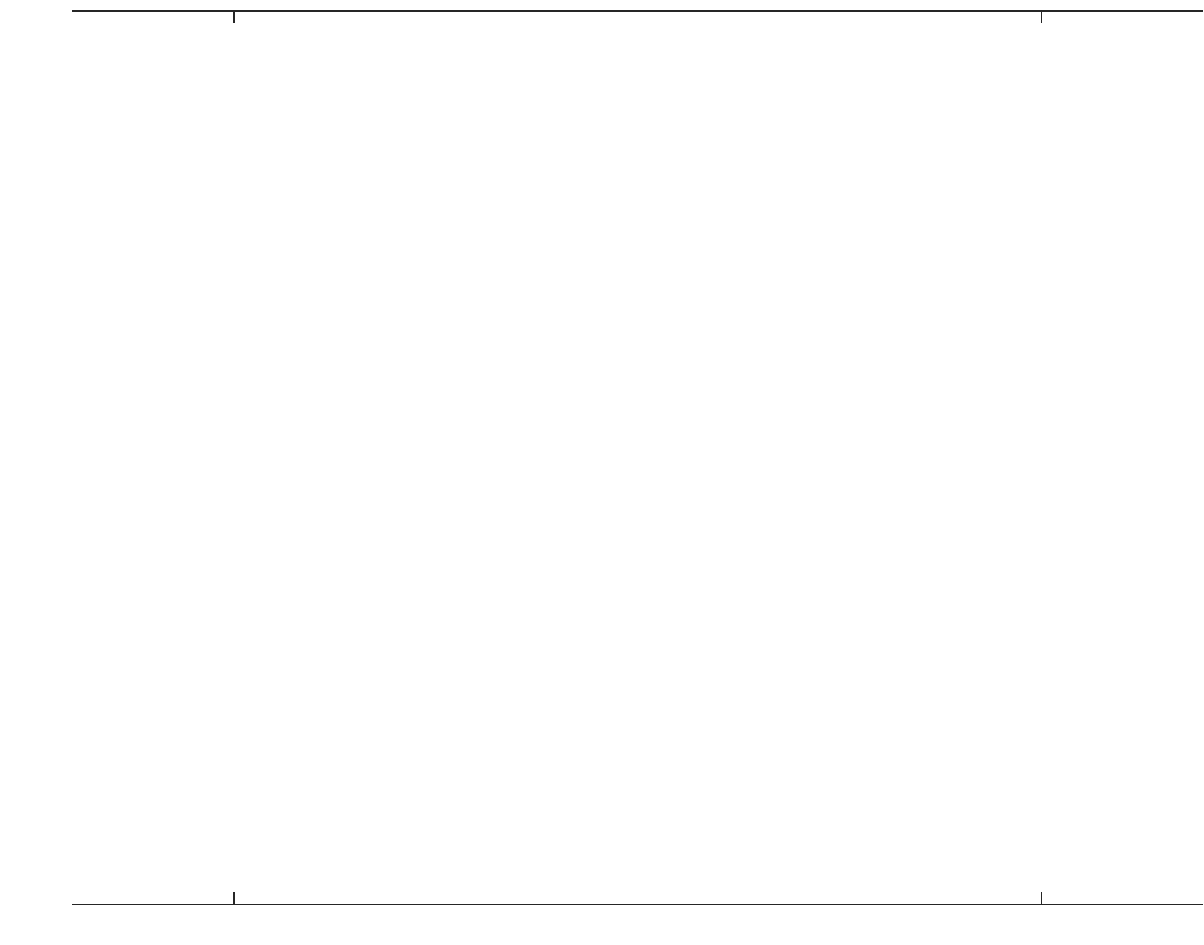}}~
	\end{center}
	\caption{Boxplots of \tabref{tab:MTRINC_aRRMSE_Outlier00} (a), \tabref{tab:MTRINC_aRRMSE_Outlier02} (b) and \tabref{tab:MTRINC_aRRMSE_Outlier04} (c). The outliers were not shown for better visualization.}
	\label{fig:boxplotsINC}
\end{figure}

As in \secref{sec:regout}, the results obtained for the incremental techniques have almost the same median (around 0.55) when trained without outliers, as we can see in \figref{fig:boxplotsINC}. When considering contaminated datasets, the median of the results increase, where IGOR-ELM achieves smaller values than IR-ELM, which is not robust to outliers. 

\begin{table}[!htb]
	\centering
	\caption{Parameter specifications of incremental methods for MTR datasets and its mean number of nodes.}

\begin{tabular}{ccccccc}
	\toprule
	\multirow{3}{*}{Dataset} & IR-ELM               & \multicolumn{5}{c}{IGOR-ELM}       \\ \cmidrule(lr){2-2} \cmidrule(lr){3-7}
	& \multirow{2}{*}{$C$} & \multirow{2}{*}{$\lambda$} & \multirow{2}{*}{$\alpha$} & \multicolumn{3}{c}{$Mean~\Ntil$}                 \\ \cmidrule(lr){5-7}
	&                      &                         &                           & No Outliers & 20\% Outliers & 40\% Outliers \\ \hline
	andro                    & $2^{20}$             & $2^{-7}$                 & $0$                       & $1000$      & $1000$        & $1000$        \\
	atp1d                    & $2^{-2}$             & $2^{-5}$                & $0$                       & $1000$      & $1000$        & $1000$        \\
	atp7d                    & $2^{-1}$             & $2^{-4}$                & $0$                       & $1000$ & $1000$ & $1000$        \\
	edm                      & $2^{9}$              & $2^{-7}$                 & $0.25$                       & $839.81$ & $794.64$ & $849.25$      \\
	enb                      & $2^{13}$             & $2^{-7}$                & $1$                       & 423.12 & 368.54 & 389.86       \\
	jura                     & $2^{0}$              & $2^{2}$                & $0$                       & $1000$      & $1000$        & $1000$        \\
	oes10                    & $2^{-1}$             & $2^{5}$                & $0$                       & $1000$      & $1000$        & $1000$        \\
	oes97                    & $2^{-2}$             & $2^{4}$                & $0$                       & $1000$      & $1000$        & $1000$        \\
	rf1                      & $2^{2}$              & $2^{-3}$                 & $0$                       & $1000$      & $1000$        & $1000$        \\
	rf2                      & $2^{3}$              & $2^{0}$                 & $0$                       & $1000$      & $1000$        & $1000$        \\
	scm1d                    & $2^{-2}$             & $2^{3}$                & $0$                       & $1000$      & $1000$        & $1000$        \\
	scm20d                   & $2^{2}$              & $2^{-3}$                & $1$                       & $999.99$ & $1000$ & $999.96$        \\
	scpf                     & $2^{20}$             & $2^{-16}$                 & $0.5$                       & $1000$ & $999.99$ & $999.94$        \\
	slump                    & $2^{20}$             & $2^{-6}$                & $1$                      & $157.40$ & $164.13$ & $159.24$       \\
	wq                       & $2^{19}$             & $2^{-19}$                & $0$                      & $1000$      & $1000$        & $1000$          \\ 
	\bottomrule
\end{tabular}

	\label{tab:MTRINC_params}
\end{table}

In our incremental approach, we use the Wilcoxon Signed Rank test \citep{Wilcoxon1945} to confirm the statistical significance of our results. We chose this test since it is more suitable to compare two methods. We consider a null hypothesis that the distribution of difference between IGOR-ELM and IR-ELM metrics has median equals to zero, against the alternate hypothesis that it is less than zero, and a significance level of 10\%.  

When considering uncontaminated datasets, the null hypothesis was not rejected, meaning that both methods are statistically equivalent. However, when the training dataset has outliers, it was rejected with p-values of 0.0014 and 0.0017, for outlier ratios of 20\% and 40\%, respectively. This means that the distribution of IGOR-ELM metrics has a inferior median than IR-ELM. Thus, the former is statistically better than the latter, since inferior metrics in regression tasks implies in better performance.

%% file: figs/b00.pdf_tex
\begingroup%
  \makeatletter%
  \providecommand\color[2][]{%
    \errmessage{(Inkscape) Color is used for the text in Inkscape, but the package 'color.sty' is not loaded}%
    \renewcommand\color[2][]{}%
  }%
  \providecommand\transparent[1]{%
    \errmessage{(Inkscape) Transparency is used (non-zero) for the text in Inkscape, but the package 'transparent.sty' is not loaded}%
    \renewcommand\transparent[1]{}%
  }%
  \providecommand\rotatebox[2]{#2}%
  \newcommand*\fsize{\dimexpr\f@size pt\relax}%
  \newcommand*\lineheight[1]{\fontsize{\fsize}{#1\fsize}\selectfont}%
  \ifx\svgwidth\undefined%
    \setlength{\unitlength}{361.92967987bp}%
    \ifx\svgscale\undefined%
      \relax%
    \else%
      \setlength{\unitlength}{\unitlength * \real{\svgscale}}%
    \fi%
  \else%
    \setlength{\unitlength}{\svgwidth}%
  \fi%
  \global\let\svgwidth\undefined%
  \global\let\svgscale\undefined%
  \makeatother%
  \begin{picture}(1,0.77482486)%
    \lineheight{1}%
    \setlength\tabcolsep{0pt}%
    \put(0,0){\includegraphics[width=\unitlength,page=1]{b00.pdf}}%
    \put(0.16161196,0.00035212){\color[rgb]{0.14901961,0.14901961,0.14901961}\makebox(0,0)[lt]{\lineheight{1.25}\smash{\begin{tabular}[t]{l}R-ELM\end{tabular}}}}%
    \put(0.37815954,0.00035212){\color[rgb]{0.14901961,0.14901961,0.14901961}\makebox(0,0)[lt]{\lineheight{1.25}\smash{\begin{tabular}[t]{l}OR-ELM\end{tabular}}}}%
    \put(0.60506829,0.00035212){\color[rgb]{0.14901961,0.14901961,0.14901961}\makebox(0,0)[lt]{\lineheight{1.25}\smash{\begin{tabular}[t]{l}GR-ELM\end{tabular}}}}%
    \put(0.82057977,0.00035212){\color[rgb]{0.14901961,0.14901961,0.14901961}\makebox(0,0)[lt]{\lineheight{1.25}\smash{\begin{tabular}[t]{l}GOR-ELM\end{tabular}}}}%
    \put(0,0){\includegraphics[width=\unitlength,page=2]{b00.pdf}}%
    \put(0.0619725,0.02901786){\color[rgb]{0.14901961,0.14901961,0.14901961}\makebox(0,0)[lt]{\lineheight{1.25}\smash{\begin{tabular}[t]{l}0\end{tabular}}}}%
    \put(0.03503357,0.11993677){\color[rgb]{0.14901961,0.14901961,0.14901961}\makebox(0,0)[lt]{\lineheight{1.25}\smash{\begin{tabular}[t]{l}0.2\end{tabular}}}}%
    \put(0.03503357,0.21085567){\color[rgb]{0.14901961,0.14901961,0.14901961}\makebox(0,0)[lt]{\lineheight{1.25}\smash{\begin{tabular}[t]{l}0.4\end{tabular}}}}%
    \put(0.03503357,0.30177458){\color[rgb]{0.14901961,0.14901961,0.14901961}\makebox(0,0)[lt]{\lineheight{1.25}\smash{\begin{tabular}[t]{l}0.6\end{tabular}}}}%
    \put(0.03503357,0.39269348){\color[rgb]{0.14901961,0.14901961,0.14901961}\makebox(0,0)[lt]{\lineheight{1.25}\smash{\begin{tabular}[t]{l}0.8\end{tabular}}}}%
    \put(0.0619725,0.48361239){\color[rgb]{0.14901961,0.14901961,0.14901961}\makebox(0,0)[lt]{\lineheight{1.25}\smash{\begin{tabular}[t]{l}1\end{tabular}}}}%
    \put(0.03503357,0.57453129){\color[rgb]{0.14901961,0.14901961,0.14901961}\makebox(0,0)[lt]{\lineheight{1.25}\smash{\begin{tabular}[t]{l}1.2\end{tabular}}}}%
    \put(0.03503357,0.6654502){\color[rgb]{0.14901961,0.14901961,0.14901961}\makebox(0,0)[lt]{\lineheight{1.25}\smash{\begin{tabular}[t]{l}1.4\end{tabular}}}}%
    \put(0.03503357,0.7563691){\color[rgb]{0.14901961,0.14901961,0.14901961}\makebox(0,0)[lt]{\lineheight{1.25}\smash{\begin{tabular}[t]{l}1.6\end{tabular}}}}%
    \put(0.01845576,0.33881602){\color[rgb]{0.14901961,0.14901961,0.14901961}\rotatebox{90}{\makebox(0,0)[lt]{\lineheight{1.25}\smash{\begin{tabular}[t]{l}aRRMSE\end{tabular}}}}}%
    \put(0,0){\includegraphics[width=\unitlength,page=3]{b00.pdf}}%
  \end{picture}%
\endgroup%

%% file: figs/b02.pdf_tex
\begingroup%
  \makeatletter%
  \providecommand\color[2][]{%
    \errmessage{(Inkscape) Color is used for the text in Inkscape, but the package 'color.sty' is not loaded}%
    \renewcommand\color[2][]{}%
  }%
  \providecommand\transparent[1]{%
    \errmessage{(Inkscape) Transparency is used (non-zero) for the text in Inkscape, but the package 'transparent.sty' is not loaded}%
    \renewcommand\transparent[1]{}%
  }%
  \providecommand\rotatebox[2]{#2}%
  \newcommand*\fsize{\dimexpr\f@size pt\relax}%
  \newcommand*\lineheight[1]{\fontsize{\fsize}{#1\fsize}\selectfont}%
  \ifx\svgwidth\undefined%
    \setlength{\unitlength}{361.92967987bp}%
    \ifx\svgscale\undefined%
      \relax%
    \else%
      \setlength{\unitlength}{\unitlength * \real{\svgscale}}%
    \fi%
  \else%
    \setlength{\unitlength}{\svgwidth}%
  \fi%
  \global\let\svgwidth\undefined%
  \global\let\svgscale\undefined%
  \makeatother%
  \begin{picture}(1,0.77482486)%
    \lineheight{1}%
    \setlength\tabcolsep{0pt}%
    \put(0,0){\includegraphics[width=\unitlength,page=1]{b02.pdf}}%
    \put(0.16161196,0.00035212){\color[rgb]{0.14901961,0.14901961,0.14901961}\makebox(0,0)[lt]{\lineheight{1.25}\smash{\begin{tabular}[t]{l}R-ELM\end{tabular}}}}%
    \put(0.37815954,0.00035212){\color[rgb]{0.14901961,0.14901961,0.14901961}\makebox(0,0)[lt]{\lineheight{1.25}\smash{\begin{tabular}[t]{l}OR-ELM\end{tabular}}}}%
    \put(0.60506829,0.00035212){\color[rgb]{0.14901961,0.14901961,0.14901961}\makebox(0,0)[lt]{\lineheight{1.25}\smash{\begin{tabular}[t]{l}GR-ELM\end{tabular}}}}%
    \put(0.82057977,0.00035212){\color[rgb]{0.14901961,0.14901961,0.14901961}\makebox(0,0)[lt]{\lineheight{1.25}\smash{\begin{tabular}[t]{l}GOR-ELM\end{tabular}}}}%
    \put(0,0){\includegraphics[width=\unitlength,page=2]{b02.pdf}}%
    \put(0.0619725,0.02901786){\color[rgb]{0.14901961,0.14901961,0.14901961}\makebox(0,0)[lt]{\lineheight{1.25}\smash{\begin{tabular}[t]{l}0\end{tabular}}}}%
    \put(0.03503357,0.1329253){\color[rgb]{0.14901961,0.14901961,0.14901961}\makebox(0,0)[lt]{\lineheight{1.25}\smash{\begin{tabular}[t]{l}0.5\end{tabular}}}}%
    \put(0.0619725,0.23683248){\color[rgb]{0.14901961,0.14901961,0.14901961}\makebox(0,0)[lt]{\lineheight{1.25}\smash{\begin{tabular}[t]{l}1\end{tabular}}}}%
    \put(0.03503357,0.34073989){\color[rgb]{0.14901961,0.14901961,0.14901961}\makebox(0,0)[lt]{\lineheight{1.25}\smash{\begin{tabular}[t]{l}1.5\end{tabular}}}}%
    \put(0.0619725,0.44464707){\color[rgb]{0.14901961,0.14901961,0.14901961}\makebox(0,0)[lt]{\lineheight{1.25}\smash{\begin{tabular}[t]{l}2\end{tabular}}}}%
    \put(0.03503357,0.54855446){\color[rgb]{0.14901961,0.14901961,0.14901961}\makebox(0,0)[lt]{\lineheight{1.25}\smash{\begin{tabular}[t]{l}2.5\end{tabular}}}}%
    \put(0.0619725,0.65246169){\color[rgb]{0.14901961,0.14901961,0.14901961}\makebox(0,0)[lt]{\lineheight{1.25}\smash{\begin{tabular}[t]{l}3\end{tabular}}}}%
    \put(0.03503357,0.7563691){\color[rgb]{0.14901961,0.14901961,0.14901961}\makebox(0,0)[lt]{\lineheight{1.25}\smash{\begin{tabular}[t]{l}3.5\end{tabular}}}}%
    \put(0.01845576,0.33881602){\color[rgb]{0.14901961,0.14901961,0.14901961}\rotatebox{90}{\makebox(0,0)[lt]{\lineheight{1.25}\smash{\begin{tabular}[t]{l}aRRMSE\end{tabular}}}}}%
    \put(0,0){\includegraphics[width=\unitlength,page=3]{b02.pdf}}%
  \end{picture}%
\endgroup%

%% file: figs/b04.pdf_tex
\begingroup%
  \makeatletter%
  \providecommand\color[2][]{%
    \errmessage{(Inkscape) Color is used for the text in Inkscape, but the package 'color.sty' is not loaded}%
    \renewcommand\color[2][]{}%
  }%
  \providecommand\transparent[1]{%
    \errmessage{(Inkscape) Transparency is used (non-zero) for the text in Inkscape, but the package 'transparent.sty' is not loaded}%
    \renewcommand\transparent[1]{}%
  }%
  \providecommand\rotatebox[2]{#2}%
  \newcommand*\fsize{\dimexpr\f@size pt\relax}%
  \newcommand*\lineheight[1]{\fontsize{\fsize}{#1\fsize}\selectfont}%
  \ifx\svgwidth\undefined%
    \setlength{\unitlength}{361.92967987bp}%
    \ifx\svgscale\undefined%
      \relax%
    \else%
      \setlength{\unitlength}{\unitlength * \real{\svgscale}}%
    \fi%
  \else%
    \setlength{\unitlength}{\svgwidth}%
  \fi%
  \global\let\svgwidth\undefined%
  \global\let\svgscale\undefined%
  \makeatother%
  \begin{picture}(1,0.77449703)%
    \lineheight{1}%
    \setlength\tabcolsep{0pt}%
    \put(0,0){\includegraphics[width=\unitlength,page=1]{b04.pdf}}%
    \put(0.16161196,0.00035212){\color[rgb]{0.14901961,0.14901961,0.14901961}\makebox(0,0)[lt]{\lineheight{1.25}\smash{\begin{tabular}[t]{l}R-ELM\end{tabular}}}}%
    \put(0.37815954,0.00035212){\color[rgb]{0.14901961,0.14901961,0.14901961}\makebox(0,0)[lt]{\lineheight{1.25}\smash{\begin{tabular}[t]{l}OR-ELM\end{tabular}}}}%
    \put(0.60506829,0.00035212){\color[rgb]{0.14901961,0.14901961,0.14901961}\makebox(0,0)[lt]{\lineheight{1.25}\smash{\begin{tabular}[t]{l}GR-ELM\end{tabular}}}}%
    \put(0.82057977,0.00035212){\color[rgb]{0.14901961,0.14901961,0.14901961}\makebox(0,0)[lt]{\lineheight{1.25}\smash{\begin{tabular}[t]{l}GOR-ELM\end{tabular}}}}%
    \put(0,0){\includegraphics[width=\unitlength,page=2]{b04.pdf}}%
    \put(0.0619725,0.02901786){\color[rgb]{0.14901961,0.14901961,0.14901961}\makebox(0,0)[lt]{\lineheight{1.25}\smash{\begin{tabular}[t]{l}0\end{tabular}}}}%
    \put(0.03503357,0.101753){\color[rgb]{0.14901961,0.14901961,0.14901961}\makebox(0,0)[lt]{\lineheight{1.25}\smash{\begin{tabular}[t]{l}0.5\end{tabular}}}}%
    \put(0.0619725,0.17448814){\color[rgb]{0.14901961,0.14901961,0.14901961}\makebox(0,0)[lt]{\lineheight{1.25}\smash{\begin{tabular}[t]{l}1\end{tabular}}}}%
    \put(0.03503357,0.24722321){\color[rgb]{0.14901961,0.14901961,0.14901961}\makebox(0,0)[lt]{\lineheight{1.25}\smash{\begin{tabular}[t]{l}1.5\end{tabular}}}}%
    \put(0.0619725,0.31995835){\color[rgb]{0.14901961,0.14901961,0.14901961}\makebox(0,0)[lt]{\lineheight{1.25}\smash{\begin{tabular}[t]{l}2\end{tabular}}}}%
    \put(0.03503357,0.39269348){\color[rgb]{0.14901961,0.14901961,0.14901961}\makebox(0,0)[lt]{\lineheight{1.25}\smash{\begin{tabular}[t]{l}2.5\end{tabular}}}}%
    \put(0.0619725,0.46542862){\color[rgb]{0.14901961,0.14901961,0.14901961}\makebox(0,0)[lt]{\lineheight{1.25}\smash{\begin{tabular}[t]{l}3\end{tabular}}}}%
    \put(0.03503357,0.53816372){\color[rgb]{0.14901961,0.14901961,0.14901961}\makebox(0,0)[lt]{\lineheight{1.25}\smash{\begin{tabular}[t]{l}3.5\end{tabular}}}}%
    \put(0.0619725,0.61089886){\color[rgb]{0.14901961,0.14901961,0.14901961}\makebox(0,0)[lt]{\lineheight{1.25}\smash{\begin{tabular}[t]{l}4\end{tabular}}}}%
    \put(0.03503357,0.68363398){\color[rgb]{0.14901961,0.14901961,0.14901961}\makebox(0,0)[lt]{\lineheight{1.25}\smash{\begin{tabular}[t]{l}4.5\end{tabular}}}}%
    \put(0.0619725,0.7563691){\color[rgb]{0.14901961,0.14901961,0.14901961}\makebox(0,0)[lt]{\lineheight{1.25}\smash{\begin{tabular}[t]{l}5\end{tabular}}}}%
    \put(0.01845576,0.33881602){\color[rgb]{0.14901961,0.14901961,0.14901961}\rotatebox{90}{\makebox(0,0)[lt]{\lineheight{1.25}\smash{\begin{tabular}[t]{l}aRRMSE\end{tabular}}}}}%
    \put(0,0){\includegraphics[width=\unitlength,page=3]{b04.pdf}}%
  \end{picture}%
\endgroup%

%% file: figs/i00.pdf_tex
\begingroup%
  \makeatletter%
  \providecommand\color[2][]{%
    \errmessage{(Inkscape) Color is used for the text in Inkscape, but the package 'color.sty' is not loaded}%
    \renewcommand\color[2][]{}%
  }%
  \providecommand\transparent[1]{%
    \errmessage{(Inkscape) Transparency is used (non-zero) for the text in Inkscape, but the package 'transparent.sty' is not loaded}%
    \renewcommand\transparent[1]{}%
  }%
  \providecommand\rotatebox[2]{#2}%
  \newcommand*\fsize{\dimexpr\f@size pt\relax}%
  \newcommand*\lineheight[1]{\fontsize{\fsize}{#1\fsize}\selectfont}%
  \ifx\svgwidth\undefined%
    \setlength{\unitlength}{354.09507751bp}%
    \ifx\svgscale\undefined%
      \relax%
    \else%
      \setlength{\unitlength}{\unitlength * \real{\svgscale}}%
    \fi%
  \else%
    \setlength{\unitlength}{\svgwidth}%
  \fi%
  \global\let\svgwidth\undefined%
  \global\let\svgscale\undefined%
  \makeatother%
  \begin{picture}(1,0.76822984)%
    \lineheight{1}%
    \setlength\tabcolsep{0pt}%
    \put(0,0){\includegraphics[width=\unitlength,page=1]{i00.pdf}}%
    \put(0.15218152,0.00033328){\color[rgb]{0.14901961,0.14901961,0.14901961}\makebox(0,0)[lt]{\lineheight{1.25}\smash{\begin{tabular}[t]{l}IGOR-ELM\end{tabular}}}}%
    \put(0.8278475,0.00033328){\color[rgb]{0.14901961,0.14901961,0.14901961}\makebox(0,0)[lt]{\lineheight{1.25}\smash{\begin{tabular}[t]{l}IR-ELM\end{tabular}}}}%
    \put(0,0){\includegraphics[width=\unitlength,page=2]{i00.pdf}}%
    \put(0.05380857,0.02457356){\color[rgb]{0.14901961,0.14901961,0.14901961}\makebox(0,0)[lt]{\lineheight{1.25}\smash{\begin{tabular}[t]{l}0\end{tabular}}}}%
    \put(0.03262782,0.26674012){\color[rgb]{0.14901961,0.14901961,0.14901961}\makebox(0,0)[lt]{\lineheight{1.25}\smash{\begin{tabular}[t]{l}0.5\end{tabular}}}}%
    \put(0.05380857,0.50890685){\color[rgb]{0.14901961,0.14901961,0.14901961}\makebox(0,0)[lt]{\lineheight{1.25}\smash{\begin{tabular}[t]{l}1\end{tabular}}}}%
    \put(0.03262782,0.75107336){\color[rgb]{0.14901961,0.14901961,0.14901961}\makebox(0,0)[lt]{\lineheight{1.25}\smash{\begin{tabular}[t]{l}1.5\end{tabular}}}}%
    \put(0.01921341,0.34228526){\color[rgb]{0.14901961,0.14901961,0.14901961}\rotatebox{90}{\makebox(0,0)[lt]{\lineheight{1.25}\smash{\begin{tabular}[t]{l}aRRMSE\end{tabular}}}}}%
    \put(0,0){\includegraphics[width=\unitlength,page=3]{i00.pdf}}%
  \end{picture}%
\endgroup%

%% file: figs/i02.pdf_tex
\begingroup%
  \makeatletter%
  \providecommand\color[2][]{%
    \errmessage{(Inkscape) Color is used for the text in Inkscape, but the package 'color.sty' is not loaded}%
    \renewcommand\color[2][]{}%
  }%
  \providecommand\transparent[1]{%
    \errmessage{(Inkscape) Transparency is used (non-zero) for the text in Inkscape, but the package 'transparent.sty' is not loaded}%
    \renewcommand\transparent[1]{}%
  }%
  \providecommand\rotatebox[2]{#2}%
  \newcommand*\fsize{\dimexpr\f@size pt\relax}%
  \newcommand*\lineheight[1]{\fontsize{\fsize}{#1\fsize}\selectfont}%
  \ifx\svgwidth\undefined%
    \setlength{\unitlength}{346.59507751bp}%
    \ifx\svgscale\undefined%
      \relax%
    \else%
      \setlength{\unitlength}{\unitlength * \real{\svgscale}}%
    \fi%
  \else%
    \setlength{\unitlength}{\svgwidth}%
  \fi%
  \global\let\svgwidth\undefined%
  \global\let\svgscale\undefined%
  \makeatother%
  \begin{picture}(1,0.78517061)%
    \lineheight{1}%
    \setlength\tabcolsep{0pt}%
    \put(0,0){\includegraphics[width=\unitlength,page=1]{i02.pdf}}%
    \put(0.13383549,0.00034049){\color[rgb]{0.14901961,0.14901961,0.14901961}\makebox(0,0)[lt]{\lineheight{1.25}\smash{\begin{tabular}[t]{l}IGOR-ELM\end{tabular}}}}%
    \put(0.82412228,0.00034049){\color[rgb]{0.14901961,0.14901961,0.14901961}\makebox(0,0)[lt]{\lineheight{1.25}\smash{\begin{tabular}[t]{l}IR-ELM\end{tabular}}}}%
    \put(0,0){\includegraphics[width=\unitlength,page=2]{i02.pdf}}%
    \put(0.03333385,0.02510531){\color[rgb]{0.14901961,0.14901961,0.14901961}\makebox(0,0)[lt]{\lineheight{1.25}\smash{\begin{tabular}[t]{l}0\end{tabular}}}}%
    \put(0.03333385,0.14880879){\color[rgb]{0.14901961,0.14901961,0.14901961}\makebox(0,0)[lt]{\lineheight{1.25}\smash{\begin{tabular}[t]{l}1\end{tabular}}}}%
    \put(0.03333385,0.27251213){\color[rgb]{0.14901961,0.14901961,0.14901961}\makebox(0,0)[lt]{\lineheight{1.25}\smash{\begin{tabular}[t]{l}2\end{tabular}}}}%
    \put(0.03333385,0.3962156){\color[rgb]{0.14901961,0.14901961,0.14901961}\makebox(0,0)[lt]{\lineheight{1.25}\smash{\begin{tabular}[t]{l}3\end{tabular}}}}%
    \put(0.03333385,0.51991911){\color[rgb]{0.14901961,0.14901961,0.14901961}\makebox(0,0)[lt]{\lineheight{1.25}\smash{\begin{tabular}[t]{l}4\end{tabular}}}}%
    \put(0.03333385,0.64362239){\color[rgb]{0.14901961,0.14901961,0.14901961}\makebox(0,0)[lt]{\lineheight{1.25}\smash{\begin{tabular}[t]{l}5\end{tabular}}}}%
    \put(0.03333385,0.7673259){\color[rgb]{0.14901961,0.14901961,0.14901961}\makebox(0,0)[lt]{\lineheight{1.25}\smash{\begin{tabular}[t]{l}6\end{tabular}}}}%
    \put(0.01962917,0.349692){\color[rgb]{0.14901961,0.14901961,0.14901961}\rotatebox{90}{\makebox(0,0)[lt]{\lineheight{1.25}\smash{\begin{tabular}[t]{l}aRRMSE\end{tabular}}}}}%
    \put(0,0){\includegraphics[width=\unitlength,page=3]{i02.pdf}}%
  \end{picture}%
\endgroup%

%% file: figs/i04.pdf_tex
\begingroup%
  \makeatletter%
  \providecommand\color[2][]{%
    \errmessage{(Inkscape) Color is used for the text in Inkscape, but the package 'color.sty' is not loaded}%
    \renewcommand\color[2][]{}%
  }%
  \providecommand\transparent[1]{%
    \errmessage{(Inkscape) Transparency is used (non-zero) for the text in Inkscape, but the package 'transparent.sty' is not loaded}%
    \renewcommand\transparent[1]{}%
  }%
  \providecommand\rotatebox[2]{#2}%
  \newcommand*\fsize{\dimexpr\f@size pt\relax}%
  \newcommand*\lineheight[1]{\fontsize{\fsize}{#1\fsize}\selectfont}%
  \ifx\svgwidth\undefined%
    \setlength{\unitlength}{346.59507751bp}%
    \ifx\svgscale\undefined%
      \relax%
    \else%
      \setlength{\unitlength}{\unitlength * \real{\svgscale}}%
    \fi%
  \else%
    \setlength{\unitlength}{\svgwidth}%
  \fi%
  \global\let\svgwidth\undefined%
  \global\let\svgscale\undefined%
  \makeatother%
  \begin{picture}(1,0.78517061)%
    \lineheight{1}%
    \setlength\tabcolsep{0pt}%
    \put(0,0){\includegraphics[width=\unitlength,page=1]{i04.pdf}}%
    \put(0.13383549,0.00034049){\color[rgb]{0.14901961,0.14901961,0.14901961}\makebox(0,0)[lt]{\lineheight{1.25}\smash{\begin{tabular}[t]{l}IGOR-ELM\end{tabular}}}}%
    \put(0.82412228,0.00034049){\color[rgb]{0.14901961,0.14901961,0.14901961}\makebox(0,0)[lt]{\lineheight{1.25}\smash{\begin{tabular}[t]{l}IR-ELM\end{tabular}}}}%
    \put(0,0){\includegraphics[width=\unitlength,page=2]{i04.pdf}}%
    \put(0.03333385,0.02510531){\color[rgb]{0.14901961,0.14901961,0.14901961}\makebox(0,0)[lt]{\lineheight{1.25}\smash{\begin{tabular}[t]{l}0\end{tabular}}}}%
    \put(0.03333385,0.14880879){\color[rgb]{0.14901961,0.14901961,0.14901961}\makebox(0,0)[lt]{\lineheight{1.25}\smash{\begin{tabular}[t]{l}1\end{tabular}}}}%
    \put(0.03333385,0.27251213){\color[rgb]{0.14901961,0.14901961,0.14901961}\makebox(0,0)[lt]{\lineheight{1.25}\smash{\begin{tabular}[t]{l}2\end{tabular}}}}%
    \put(0.03333385,0.3962156){\color[rgb]{0.14901961,0.14901961,0.14901961}\makebox(0,0)[lt]{\lineheight{1.25}\smash{\begin{tabular}[t]{l}3\end{tabular}}}}%
    \put(0.03333385,0.51991911){\color[rgb]{0.14901961,0.14901961,0.14901961}\makebox(0,0)[lt]{\lineheight{1.25}\smash{\begin{tabular}[t]{l}4\end{tabular}}}}%
    \put(0.03333385,0.64362239){\color[rgb]{0.14901961,0.14901961,0.14901961}\makebox(0,0)[lt]{\lineheight{1.25}\smash{\begin{tabular}[t]{l}5\end{tabular}}}}%
    \put(0.03333385,0.7673259){\color[rgb]{0.14901961,0.14901961,0.14901961}\makebox(0,0)[lt]{\lineheight{1.25}\smash{\begin{tabular}[t]{l}6\end{tabular}}}}%
    \put(0.01962917,0.349692){\color[rgb]{0.14901961,0.14901961,0.14901961}\rotatebox{90}{\makebox(0,0)[lt]{\lineheight{1.25}\smash{\begin{tabular}[t]{l}aRRMSE\end{tabular}}}}}%
    \put(0,0){\includegraphics[width=\unitlength,page=3]{i04.pdf}}%
  \end{picture}%
\endgroup%

%% file: sec/conclusions.tex
\section{Conclusions}\label{sec:Concl}

In this paper, we proposed GOR-ELM and its incremental version (IGOR-ELM), which extends the OR-ELM and GR-ELM algorithms to deal with multi-target regression problems. Instead of considering the Frobenius norm in the model error, we use the $\ell_{2,1}$ norm, which is more robust to outliers. For the proposed method, OR-ELM is a particular case of GOR-ELM, when the output has one dimension and when we set $\alpha = 0$ and $\lambda = 1$.

When we consider the non-incremental algorithms, our experiments showed that the proposed method achieved similar values of aRRMSE to those obtained by the other techniques, even when the dataset is not contaminated. However, its training stage was slightly slower. As expected, when we randomly inserted outliers in the training stage (20\% and 40\% of the samples), GOR-ELM usually showed better performance when compared to the other techniques. As the Friedman and Nemenyi statistical tests showed, GOR-ELM is better than the compared techniques in the presence of outliers.

When we consider the number of nodes, our experiments showed that GOR-ELM was, in most cases, not capable of reducing the number of its nodes without compromising the model error. Thus, if the desired model error was not achieved, IGOR-ELM can be used to increase the node number of the network. We compared this technique with IR-ELM, which can add nodes to a SLFN trained using R-ELM. When considering the original datasets, similar results were obtained. However, when the datasets were contaminated, IGOR-ELM showed better performance than IR-ELM in most cases. The Wilcoxon signed rank test showed that IGOR-ELM was better than IR-ELM in tasks with outliers.

For future work, we can consider using other types of hidden nodes and activation functions, since we only tested the sigmoidal additive ones. We can also improve the network training time using GPUs or more efficient linear algebra libraries. Additionaly, we can also consider training GOR-ELM in a distributed approach, since ADMM can be trained using multiple processors, where each one handles part of a dataset.